\documentclass[10pt,twocolumn,letterpaper]{article}

\usepackage[pagenumbers]{iccv} 

\definecolor{iccvblue}{rgb}{0.21,0.49,0.74}
\usepackage[pagebackref,breaklinks,colorlinks,allcolors=iccvblue]{hyperref}

\usepackage{multirow}
\usepackage{multicol}
\usepackage{dirtytalk}
\usepackage{breqn}
\usepackage{pifont}
\newcommand{\cmark}{\ding{51}}%
\usepackage{algorithm}
\usepackage{algorithmic}

\def\paperID{5981} 
\def\confName{ICCV}
\def\confYear{2025}

\title{M2SFormer: Multi-Spectral and Multi-Scale Attention with Edge-Aware Difficulty Guidance for Image Forgery Localization}

\author{Ju-Hyeon Nam$^{1}$ \qquad Dong-Hyun Moon$^{1}$ \qquad Sang-Chul Lee$^{1, 2}$ \\
Department of Electrical and Computer Engineering, Inha University$^{1}$,  DeepCardio$^{2}$\\
{\tt\small \{jhnam0514, 12201723\}@inha.edu \qquad sclee@\{inha.ac.kr, deepcardio.com\}}
}

\begin{document}
\maketitle
\begin{abstract}
Image editing techniques have rapidly advanced, facilitating both innovative use cases and malicious manipulation of digital images. Deep learning-based methods have recently achieved high accuracy in pixel-level forgery localization, yet they frequently struggle with computational overhead and limited representation power, particularly for subtle or complex tampering. In this paper, we propose \textbf{M2SFormer}, a novel Transformer encoder-based framework designed to overcome these challenges. Unlike approaches that process spatial and frequency cues separately, M2SFormer unifies multi-frequency and multi-scale attentions in the skip connection, harnessing global context to better capture diverse forgery artifacts. Additionally, our framework addresses the loss of fine detail during upsampling by utilizing a global prior map—a curvature metric indicating the difficulty of forgery localization—which then guides a difficulty-guided attention module to preserve subtle manipulations more effectively. Extensive experiments on multiple benchmark datasets demonstrate that M2SFormer outperforms existing state-of-the-art models, offering superior generalization in detecting and localizing forgeries across unseen domains.
\end{abstract}   
\section{Introduction}
\label{sec:intro}

\textit{Can we trust the information in the media?} With the rapid advancement of image editing techniques, the ability to manipulate digital images has become more sophisticated, enabling both creative applications and malicious modifications \cite{kawar2023imagic, yang2023paint, liu2024referring, lee2024diffusion}. This has raised significant concerns, as forged images can lead to misinformation, legal conflicts, and public distrust, contributing to societal instability \cite{vaccari2020deepfakes, mubarak2023survey, ahmed2023perception}. Consequently, image forgery detection has gained substantial attention, with early approaches leveraging traditional methods based on primitive hand-crafted features such as JPEG compression traces \cite{popescu2004statistical, yang2020clustering}, sensor pattern noise (SPN) \cite{chierchia2014bayesian, korus2016multi}, and interpolation patterns of Color Filter Array (CFA) \cite{dirik2009image, ferrara2012image}. However, these methods primarily focus on image-level detection (\textit{binary classification}) and have limited generalization ability to unseen forgery types \cite{wu2019mantra, hu2020span}.

Due to the rise of the deep learning era, pixel-level forgery localization (\textit{binary segmentation}) has become more achievable, with models like MantraNet \cite{wu2019mantra} and SPAN \cite{hu2020span} showing strong performance in automatic forgery trace extraction. However, despite their success, these methods are often computationally expensive and suffer from low representation power, which hinders their ability to generalize well to unseen types of forgeries \cite{zhang2021multi, chen2021image}. Recently, the advent of attention mechanisms \cite{hu2018squeeze, woo2018cbam, dai2021attentional} has alleviated some of these challenges, improving both the efficiency and performance of forgery localization models \cite{zhang2021multi, hao2021transforensics, liu2022pscc} based on UNet-like encoder-decoder architecture. Nevertheless, these methods still struggle with distinguishing between authentic and forged images in cases where the forgery is subtle or closely resembles the original content, resulting in decreased performance in certain real-world scenarios \cite{gu2022fbi, xu2024image, liu2024attention}.

Recently, frequency domain approaches are actively proposed across diverse fields, including image classification \cite{chen2019drop, li2020wavelet, qin2021fcanet, zhang2023wcanet}, semantic segmentation \cite{yang2020fda, azad2021deep, li2022wavesnet}, and object detection \cite{zhong2022detecting, lin2023frequency}, to improve the generalization ability to unseen domains and robustness to various corruptions. Additionally, analyzing signals in the frequency domain often proves more effective at revealing subtle manipulation artifacts than relying on spatial domain cues for forgery localization \cite{qian2020thinking, liu2021spatial, luo2021generalizing, gu2022fbi, guo2023hierarchical, xu2024image}. However, existing approaches typically employ spatial and frequency domains separately, without integrating them into a unified framework due to efficiency concerns. As a result, no unified attention mechanism for forgery localization has been proposed to jointly leverage both spatial and frequency information, largely due to concerns about the computational efficiency of such integration. This leads to a key research question: \textit{“how can subtle forgery features be effectively captured while efficiently integrating spatial and frequency attention?”}

To answer abovementioned question, we propose a novel frequency-spatial unified attention mechanism called \textit{Multi-Spectral and Multi-Spatial (M2S) attention block} in the skip connection for forgery localization. Our novel M2S attention block consists of two main attention mechanisms in different domains: \textit{frequency} and \textit{spatial} domains. In the frequency domain, we produce a channel-wise attention map using the basis images of the 2D Discrete Cosine Transform (2D DCT) \cite{ahmed1974discrete}. By selectively weighting the most relevant frequency components, this approach highlights crucial spectral information while preserving essential spatial context from a multi-spectral perspective. In the spatial domain, we employ a SIFT \cite{lowe2004distinctive}-inspired feature pyramid to capture subtle yet anomalous boundary cues arising from cut-and-paste operations, applying spatial attention at each pyramid level to highlight these forgery indicators effectively. Additionally, we quantitatively measure the “\textit{difficulty}” of each input sample and develop a decoder that integrates \textit{Difficulty-guided Attention (DGA)}—which converts the difficulty level into a textual representation and applies channel-wise attention—with a Transformer block, thereby enhancing the model’s ability to handle challenging samples more effectively. By integrating the \textit{M2S attention block} with a \textit{Edge-Aware DGA-based Transformer decoder}, we propose a Transformer-based architecture, called \textbf{M2SFormer}, a new forgery localization model that fully leverages global dependencies and consistently captures forged masks across various manipulation types, shapes, and sizes. Extensive experiments on multiple public benchmark datasets demonstrate that M2SFormer outperforms existing models, significantly improving generalization performance in forgery localization across unseen domains. Additionally, we also provide the impactness in the field of artificial intelligence of M2SFormer in Appendix (Section \ref{sec:broader_impact_in_artificial_intelligence}). The contributions of this paper can be summarized as follows: \vspace{-0.25cm}

\begin{itemize}
    \item We propose a novel Transformer-based forgery localization framework, called \textbf{M2SFormer}, that that efficiently integrates \textit{M2S attention block} with a \textit{Edge-Aware DGA-based Transformer decoder}, enhancing forgery localization performance compared to existing methods.

    \item The integration of multi-spectral and multi-scale attention in the skip connection to better capture forgery artifacts. Additionally, a global prior map quantifies the forgery localization difficulty, guiding a difficulty-guided attention module after upsampling to preserve fine details in challenging regions.

    \item Extensive experiments on multiple benchmark datasets demonstrate that M2SFormer outperforms existing models, significantly improving generalization performance in forgery localization across unseen domains.
\end{itemize}
\begin{figure*}
    \centering
    \includegraphics[width=\textwidth]{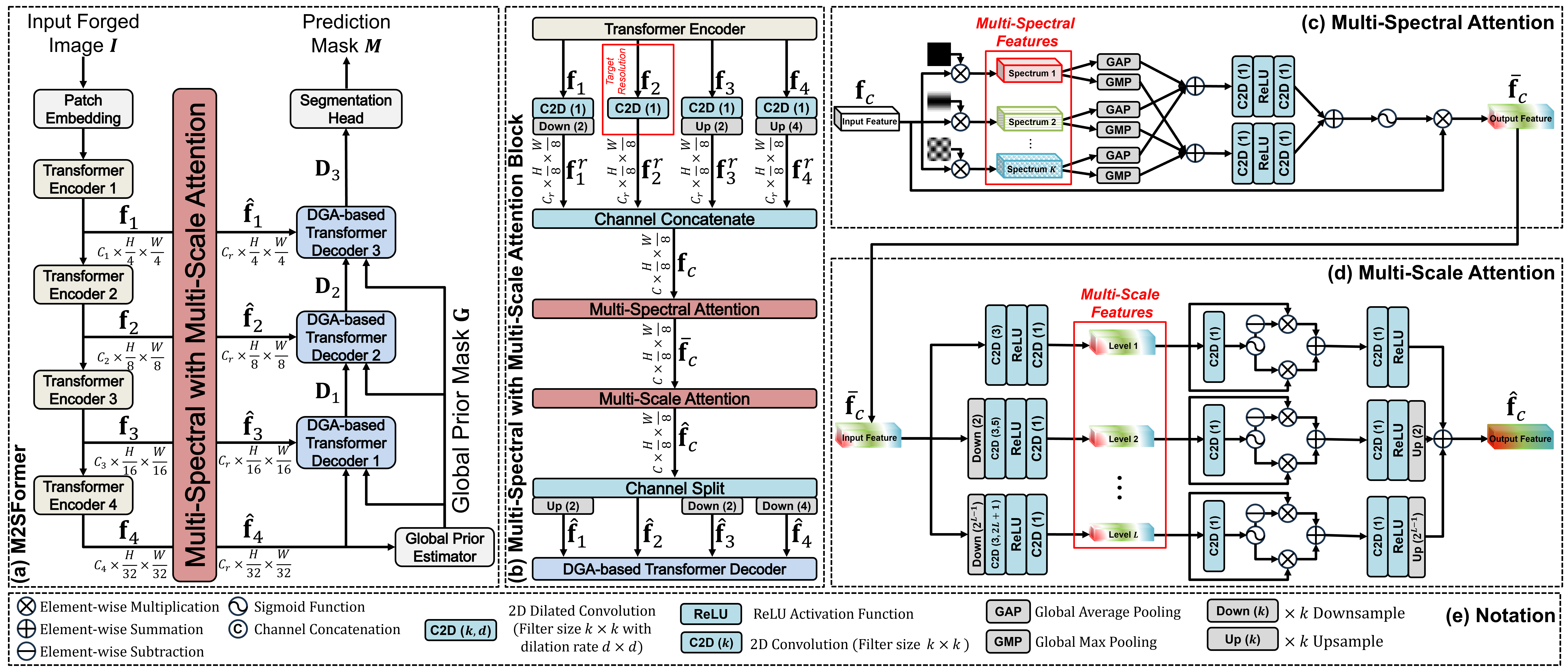} \vspace{-0.75cm}
    \caption{(a) The overall architecture of the proposed \textbf{M2SFormer} mainly comprises \textit{M2S Attention Module (Fig. \ref{fig:M2SFormer} (b))} and \textit{TGA-based Transformer Decoder (Fig. \ref{fig:dga_based_transformer_decoder})}. (b) M2S attention block. (c) Multi-Spectral attention block. (d) Multi-Scale attention block. (e) Notation description used in this paper.} \vspace{-0.75cm}
    \label{fig:M2SFormer}
\end{figure*}

\section{Related Works}
Traditional forgery detection methods, relying on JPEG compression traces \cite{popescu2004statistical, yang2020clustering}, sensor pattern noise (SPN) \cite{chierchia2014bayesian, korus2016multi}, and Color Filter Array (CFA) \cite{dirik2009image, ferrara2012image} interpolation patterns, often fail to generalize to new forged types. 

\textit{1) CNN-based approaches:} Recent advances in deep learning have addressed these limitations, enabling more robust pixel-level detection accuracy across diverse forged types. For example, RGB-Noise \cite{zhou2018learning} demonstrated robust detection across diverse forgery types by extracting noise input through an SRM filter, while MantraNet \cite{wu2019mantra} and SPAN \cite{hu2020span} leveraged CNN-based, no-pooling architectures for pixel-level localization. However, these methods typically provide only bounding-box predictions, impose high computational overhead, or require additional fine-tuning on external datasets, which can limit their practical applicability. To address this limitation, RRUNet \cite{bi2019rru} introduced a feedback learning mechanism based on a UNet-like encoder–decoder architecture. Additionally, MT-SENet \cite{zhang2021multi} incorporated SE blocks \cite{hu2018squeeze} and multi-task learning to more accurately delineate forged boundaries. Additionally, PSCCNet \cite{liu2022pscc}, HDFNet \cite{han2024hdf}, and PIMNet \cite{bai2025pim} employed a multi-scale attention structure that adaptively aggregates features at varying resolutions. 

\textit{2) Transformer-based approaches:} However, neither methods still consider the global dependencies witin the forged images, limiting their capacity to capture large-scale contextual cues for pixel-level localization. To this end, TransForensic \cite{hao2021transforensics} effectively utilized global dependency by leveraging the self-attention mechanism of Vision Transformer (ViT) \cite{dosovitskiy2021an}. 

\textit{3) Frequency-based approches:} Meanwhile, leveraging frequency-domain feature extraction has proven highly effective in enhancing cross-domain generalization. In the field of forgery localization, ObjectFormer \cite{wang2022objectformer} and FBINet \cite{gu2022fbi} utilize the 2D DCT to capture the frequency response of the input image, thereby revealing subtle tampering clues that may go undetected in purely spatial representations. Similarly, DNet \cite{yang2024d}, AFENet \cite{xu2024image}, and EITLNet \cite{guo2024effective} improve generalization performance by incorporating a Haar-based wavelet transformation, 2D discrete Fourier transform, and a high-pass noise filter with RGB input images, respectively. 

\textit{4) Our approach (\textbf{M2SFormer}):} However, most of these methods rely on training with large-scale synthetic datasets to ensure cross-domain generalizability and subsequently require fine-tuning when new external datasets become available. In contrast, the proposed \textbf{M2SFormer} leverages global dependencies by integrating the \textit{M2S attention block} and a \textit{Edge-Aware DGA-based Transformer decoder}, thereby achieving high performance on various external test datasets \textbf{without requiring additional training}. Although \cite{huh2018fighting} proposed a self-consistency model for localizing forged regions by leveraging metadata containing textual information, it inevitably depends on the presence of text input. In contrast to previous approaches, the \textit{Edge-Aware DGA-based Transformer decoder} quantifies each sample’s difficulty through a self-difficulty measurement, then automatically generates text based on this metric to allocate varying levels of attention to “\textit{difficult}” and “\textit{easy}” samples. Because our text generation process does not require external metadata, we eliminate any dependence on textual input while ensuring that more focused attention is devoted to challenging regions. \vspace{-0.15cm}

\section{Method} \vspace{-0.15cm}

\subsection{Encoder and Decoder in M2SFormer} \vspace{-0.15cm}
\label{ssec:encoder_and_decoder_in_m2sformer}

We used the ImageNet-1K \cite{russakovsky2015imagenet} pre-trained Pyramid Vision Transformer v2 (PVT-v2) \cite{wang2022pvt} as the backbone encoder. We note that PVT-v2 can efficiently extract multi-scale feature maps with a pyramid structure, and is utilized in various dense prediction tasks such as semantic segmentation and object detection than ViT \cite{dosovitskiy2021an} or Swin Transformer \cite{liu2021swin}. We utilized the same encoder architecture as the decoder to fully leverage global dependency. Although the main experimental results are presented using PVT-v2, we also include additional evaluations with various CNN and Transformer backbones in the Appendix, demonstrating the proposed approach’s versatility and robustness across different architectures (Tab. \ref{tab:ablation_backbone_networks}). \vspace{-0.15cm}

\subsection{Multi-Spectral with Multi-Scale Attention Module} \vspace{-0.15cm}
\label{ssec_m2s_attention}
\noindent \textit{Motivation:} Recent studies suggest that the human visual system (HVS) utilizes multiple frequency bands to detect subtle artifacts in forged images, highlighting the importance of multi-spectral attention in forgery localization \cite{redi2011digital}. At the same time, leveraging multiple spatial scales is crucial for capturing diverse forgery patterns of varying sizes, a principle supported by the classic feature extraction method SIFT \cite{lowe2004distinctive}. Motivated by the importance of both frequency-domain cues and scale-invariant features, we integrate multi-spectral and multi-scale attention to improve robustness and accuracy in forgery localization. This fusion allows the model to effectively capture subtle manipulation artifacts across various spatial scales while leveraging the distinct advantages offered by frequency-based representations. The proposed multi-spectral with multi-scale (M2S) attention module can be divided into four steps: \textit{1) Feature Preprocessing}, \textit{2) Multi-Spectral Attention}, \textit{3) Multi-Scale Attention}, and \textit{4) Feature Postprocessing}.

\noindent \textbf{Feature Preprocessing.} Let $\mathbf{f}_{i} \in \mathbb{R}^{C_{i} \times \frac{H}{2^{i + 1}} \times \frac{W}{2^{i + 1}}}$ denote the feature maps produced by the $i$-th encoder stage, where $i = 1, 2, 3, 4$ and $(H, W)$ is the resolution of the input image. Because the number of skip connection channels $C^{e}_{i}$ varies across stages and can significantly affect the decoder’s complexity, we apply a $1 \times 1$ convolution to reduce the channels to a uniform channel $C_{r}$. Next, each feature map is resized to the target resolution $(H_{t}, W_{t})$. Formally, \vspace{-0.15cm}
\begin{equation}
    \mathbf{f}^{'}_{i} = \textbf{Resize}_{(H_{t}, W_{t})} (\textbf{C2D}_{1 \times 1} (\mathbf{f}_{i})) \in \mathbb{R}^{C_{r} \times H_{t} \times W_{t}},
\end{equation}

\noindent where $\textbf{C2D}_{k \times k} ( \cdot )$ denotes the $1 \times 1$ convolution used for channel reduction, and $\textbf{Resize}_{(H_{t}, W_{t})} (\cdot)$ performs bilinear interpolation if the target resolution differs from the original. If the original resolution matches, no resizing is performed (marked as \say{\textit{Target Resolution}} in Figure \ref{fig:M2SFormer} (b)). Once the feature maps $\mathbf{f}^{'}_{i}$ for $i = 1, 2, 3, 4$ have been uniformly resized, they are concatenated along the channel dimension as $\mathbf{f}_{c} = \left[ \mathbf{f}^{'}_{1}, \mathbf{f}^{'}_{2}, \mathbf{f}^{'}_{3}, \mathbf{f}^{'}_{4} \right] \in \mathbb{R}^{4C_{r} \times H_{t} \times W_{t}}$ where $\left[ \cdots \right]$ denotes channel-wise concatenation. For convenience, let $C = 4C_{r}$. By following these steps, we obtain a cross-scale feature map $\mathbf{f}_{c}$ that consolidates multi-scale information into a consistent dimensionality.

\noindent \textbf{Multi-Spectral Attention.} Recently, 2D DCT has been widely employed to enhance the generalization ability of deep learning models across various domains \cite{qin2021fcanet, gu2022fbi, sang2022multi, nam2024modality}. By transforming an image or feature map into a weighted sum of cosine functions at different frequencies, 2D DCT enables more effective frequency-domain analysis. Notably, FBINet \cite{gu2022fbi} and ObjectFormer \cite{wang2022objectformer} leverage 2D DCT to reveal subtle forged traces hidden in the spatial domain and use these as discriminative features. However, applying 2D DCT directly to input images can be computationally expensive, particularly in multi-modal training schemes \cite{gu2022fbi} or dual-encoder architectures \cite{wang2022objectformer}. To address this, we propose aggregating coarse-to-fine features rather than working at the input-image level, thus making more efficient use of the frequency domain and reducing the overall computational cost. 

By leveraging the 2D DCT basis images $\mathbf{D}$, the cross-scale feature map $\mathbf{f}_{c} \in \mathbb{R}^{C \times H_{t} \times W_{t}}$ obtained from \textit{1) Feature Preprocessing} can be characterized as follows: \vspace{-0.15cm}
\begin{equation}
    \mathbf{f}^{k}_{c} = \sum_{h = 0}^{H_{t} - 1} \sum_{w = 0}^{W_{t} - 1} \left( \mathbf{f}_{c} \right)_{:, h, w} \mathbf{D}^{u_{k}, v_{k}}_{h, w},
\end{equation}

\noindent where $(u_{k}, v_{k})$ is the frequency indices corresponding to $\mathbf{f}^{k}_{c}$ and 2D DCT basis images are defined as $\mathbf{D}^{u_{k}, v_{k}}_{h, w} = \cos ( \frac{\pi h}{H_{s}} (u_{k} + \frac{1}{2}) ) \cos( \frac{\pi w}{W_{s}} (v_{k} + \frac{1}{2}) )$ with a top-$K$ selection strategy \cite{qin2021fcanet}. Next, we apply Global Average Pooling (GAP) and Global Max Pooling (GMP) to each $\mathbf{f}^{k}_{c}$ to produce $\mathbf{Z}^{k}_{\text{avg}}$ and $\mathbf{Z}^{k}_{\text{max}}$, respectively. These pooled features are then passed through a statistical aggregation block, consisting of $1 \times 1$ convolution layers with a ReLU activation, to generate a channel attention map $\mathbf{M}^{\text{channel}} \in \mathbb{R}^{C}$ as follows: \vspace{-0.15cm}
\begin{equation}
    \mathbf{M}^{\text{spectral}} = \sigma \left( \sum_{d \in \{ \text{avg}, \text{max} \}} \sum_{k = 1}^{K} \textbf{C2D}_{1 \times 1} (\delta (\textbf{C2D}_{1 \times 1}(\mathbf{f}^{k}_{c}))) \right)
\end{equation}

\noindent where $\delta (\cdot)$ and $\sigma (\cdot)$ denote ReLU and Sigmoid activation functions, respectively. Finally, we recalibrate the channel value of the cross-scale feature map $\mathbf{f}_{c}$ using $\mathbf{M}^{\text{spectral}}$ as $\overline{\mathbf{f}}_{c} = \mathbf{f}_{c} \times \mathbf{M}^{\text{channel}} \in \mathbb{R}^{C \times H_{t} \times W_{t}}$.

Although many models \cite{qin2021fcanet, sang2022multi, nam2024modality} use multi-spectral attention mechanisms, our method differs by considering the entire feature map channels and employing a convolution-based statistical aggregation block. This design not only captures richer inter-channel interactions but also offers improved parameter efficiency and flexibility.

\noindent \textbf{Multi-Scale Attention.} To capture discriminative forgery boundary cues at various spatial scales, we construct a multi-scale feature pyramid by explicitly downsampling the channel-recalibrated feature map, inspired by SIFT \cite{lowe2004distinctive}, and then apply a spatial attention to each pyramid level. However, this multi-scale approach can produce up to $L \times C$ channels, causing high memory consumption where $L$ is the number of pyramid levels. To address this issue, we first apply a  $3 \times 3$ dilated convolution to the downscaled feature map with the ReLU activation, and finally use a $1 \times 1$ convolution to reduce the channels and preserve the spatial information for each feature map as follows: \vspace{-0.15cm}
\begin{equation}
\overline{\mathbf{f}}^{l}_{c} = \textbf{C2D}_{1\times 1} \left( \textbf{DC2D}_{3 \times 3}^{2l + 1} \left( \textbf{Down}_{l} \left( \overline{\mathbf{f}}_{c} \right) \right) \right),
\end{equation} 

\noindent where $\textbf{DC2D}_{k \times k}^{2l + 1} (\cdot)$, and $\textbf{Down}_{l} (\cdot)$ denote 2D dilated convolution with a kernel size of $k \times k$ and dilation rate of $(2l + 1) \times (2l + 1)$, and downsampling with scale factor $2^{l - 1}$, respectively. Additionally, we denote the number of channels, height, and width at the $l$-th pyramid level as $C_{l} = \text{max}(\frac{C}{\gamma^{l - 1}}, C_{\text{min}}), H_{l} = \text{max}(\frac{H}{2^{l - 1}}, H_{\text{min}})$, and $W_{l} = \text{max}(\frac{W}{2^{l - 1}}, W_{\text{min}})$, respectively. Inspired by previous spatial attention methods, we introduce two learnable parameters ($\alpha^{l}_{i}$ and $\beta^{l}_{i}$) for each level of the feature pyramid. These parameters regulate the information flow between the foreground and background, allowing more flexible control over the decomposed feature maps as follows: \vspace{-0.15cm}
\begin{equation}
    \hat{\mathbf{f}}^{l}_{i} = \textbf{C2D}_{3 \times 3} \left( \alpha^{l}_{i} (\overline{\mathbf{f}}^{l}_{i} \times \mathbf{F}^{l}_{i}) + \beta^{l}_{i} (\overline{\mathbf{f}}^{l}_{i} \times \mathbf{B}^{l}_{i})  \right),
\end{equation}

\noindent where $\mathbf{F}^{l}_{i} = \sigma ( \textbf{C2D}_{1 \times 1} (\overline{\mathbf{f}}^{l}_{i}) )$ is a foreground attention map. Accordingly, the background attention map can be derived via elementary-wise subtraction between $\mathbf{F}^{l}_{i}$ and a matrix filled with one, that is, $\mathbf{B}^{l}_{i} = 1 - \mathbf{F}^{l}_{i}$. At that point, we restore the number channel at the $l$-th scale branch $C_{l}$ in to $C$. Finally, to sum each refined feature, we apply upsampling to match the resolutions. For more stable training, we apply residual connection after the spatially refined feature map as $\hat{\mathbf{f}}_{c} = \mathbf{f}_{c} + \sum_{l = 1}^{L} \textbf{Up}_{l} ( \hat{\mathbf{f}}^{l}_{i} )$.

\noindent \textbf{Feature Postprocessing.} We first divided $\hat{\mathbf{f}}_{c}$ along the channel dimension with an equal number of channels $C_{r}$. Subsequently, each feature map was resized and applied residual connection between the original feature map to enhance the training stability for $i = 1, 2, 3, 4$ as $\hat{\mathbf{f}}_{i} = \textbf{Resize}_{(\frac{H}{2^{i + 1}}, \frac{W}{2^{i + 1}})} \left( (\hat{\mathbf{f}}_{c})_{C_{r} \cdot (i - 1):C_{r} \cdot i} \right)$. Finally, each refined feature map $\hat{\mathbf{f}}_{i}$ is fed into the $i$-th decoder stage to predict the forged mask. \vspace{-0.15cm}

\begin{figure}
    \centering
    \includegraphics[width=0.48\textwidth]{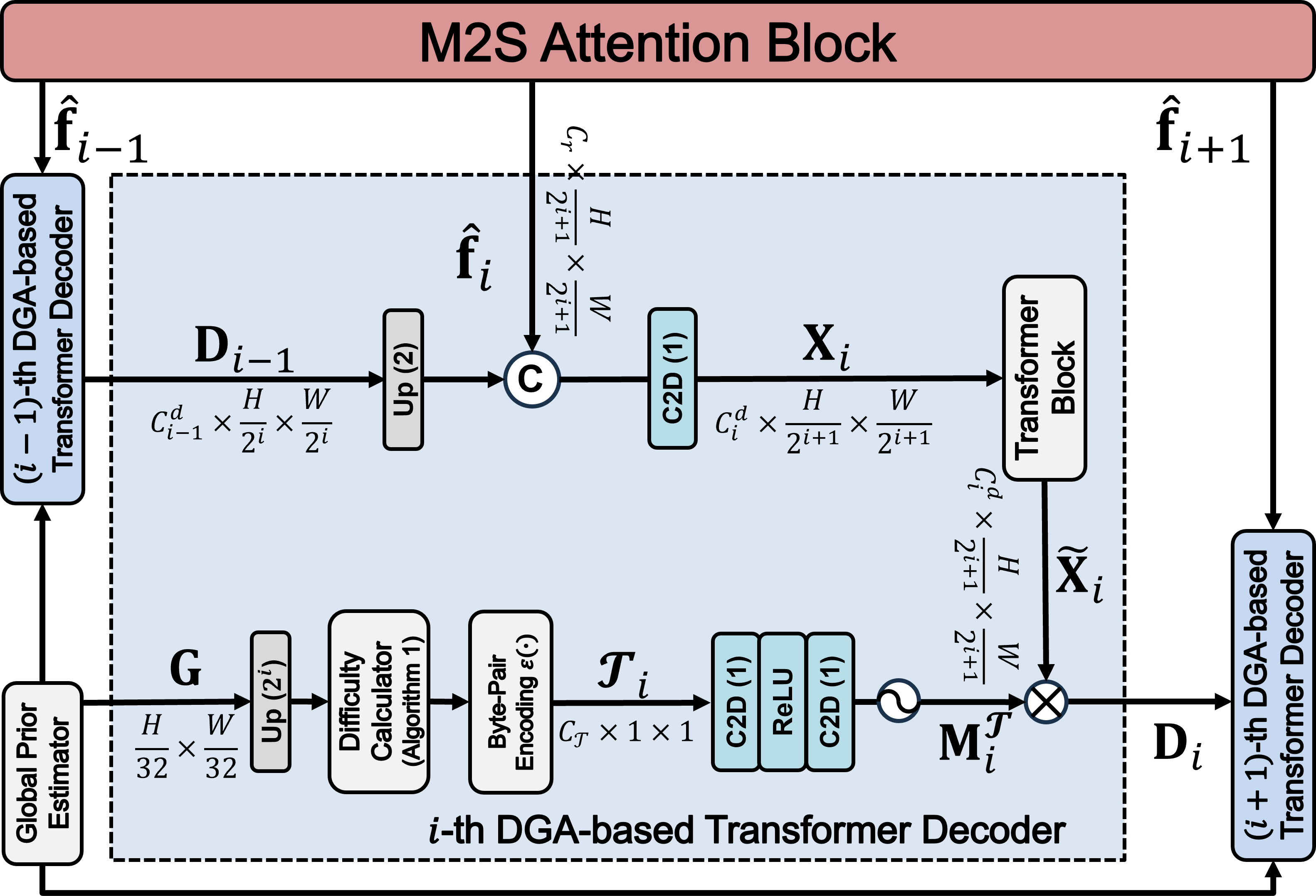} \vspace{-0.75cm}
    \caption{The overall block diagram of the Difficulty-guided Attention (DGA)-based Transformer Decoder at $i$-th stage.} \vspace{-0.5cm}
    \label{fig:dga_based_transformer_decoder}
\end{figure}

\subsection{Edge-Aware DGA-based Transformer Decoder} \vspace{-0.15cm}
\noindent \textit{Motivation:} Curvature is a key indicator of shape complexity, influencing how the human visual system allocates attention. Low-curvature regions are easily processed, while high-curvature areas demand greater perceptual resources. Biomedical studies \cite{attneave1954some, koenderink1984structure, yue2020curvature} show that curvature-based analysis helps detect anomalies, while neurophysiological research highlights its role in guiding visual focus. By using curvature to classify samples as “easy” or “hard,” a forgery localization model can refine its attention on intricate regions, improving localization accuracy. As described in Section \ref{ssec:encoder_and_decoder_in_m2sformer}, we used the same architecture in the decoder as in the encoder Transformer architecture. The only difference in our proposed Edge-Aware DGA-based Transformer decoder is adding \textit{Edge-Aware Difficulty Calculator (Algorithm \ref{alg_DLC})} and \textit{Difficulty-guided Attention (DGA, Fig. \ref{fig:dga_based_transformer_decoder})}.

\noindent \textbf{Edge-Aware Difficulty Calculator.} First, we produce a global prior map $\mathbf{G} =  \sigma(\textbf{Conv2D}_{1} (\hat{\mathbf{f}}_{4})) \in \mathbb{R}^{\frac{H}{32} \times \frac{W}{32}}$ from the deepest feature map $\hat{\mathbf{f}}_{4} \in \mathbb{R}^{C_{r} \times \frac{H}{32} \times \frac{W}{32}}$, which contains the most semantically rich information regarding forgery cues. Additionally, the deepest feature map also aggregates forgery-related evidence from earlier layers by M2S attention block (Section \ref{ssec_m2s_attention}). Consequently, the global prior map $\mathbf{G}$ can serve as a powerful guide for detecting forged regions. Then, we compute the \textit{second-order derivative based curvature} of the global prior map to quantify its difficulty level due to its simplicity and efficiency than Hessian or Gaussian curvature estimators. More specifically, we extract the first ($\textbf{G}_{x}, 
\textbf{G}_{y}$) and second ($\textbf{G}_{xx}, \textbf{G}_{xy}, \textbf{G}_{yy}$)-order derivative of $\mathbf{G}$ by an anisotropic Sobel edge detection filter \cite{kanopoulos1988design}. Finally, we can calculate the curvature map $\kappa_{i} \in \mathbb{R}^{\frac{H}{2^{i}} \times \frac{W}{2^{i}}}$ of $\mathbf{G}$ as follows: \vspace{-0.15cm}
\begin{equation}
    \kappa_{i} = \frac{\textbf{G}_{x}^{2}\textbf{G}_{yy} - 2\textbf{G}_{x}\textbf{G}_{y} + \textbf{G}_{y}^{2}\textbf{G}_{xx}}{(\textbf{G}_{x}^{2} + \textbf{G}_{y}^{2})^{1.5}}
\end{equation}

However, we noted that if taking simply the average value of the curvature map $\kappa_{i}$ over the entire image, the result will be misleadingly low, as large uniform areas of zero curvature dominate in the global prior map $\mathbf{G}$. To address this issue, we compute the average curvature only within edge regions—where non-zero curvature actually occurs—thus yielding a more representative measure of local structural changes as $\textbf{s} = \sigma(\sum (\kappa \otimes \textbf{E}) / \sum \textbf{E})$ where $\mathbf{E} = \sqrt{\textbf{G}_{x}^{2} + \textbf{G}_{y}^{2}}$ is the edge of the global prior map. Finally, if this computed value $\textbf{s}$ exceeds a specified threshold, we label the difficulty level with the text descriptor “\textit{hard}” or “\textit{easy},” depending on which side of the threshold $\tau$ it falls on. More detailed description is available in Algorithm \ref{alg_DLC}.

\noindent \textbf{Difficulty-guided Attention.} We first transform the textual difficulty level $\mathbf{t}$ (e.g., “\textit{hard}” or “\textit{easy}”) , derived from the Edge-Centric Difficulty Calculator, into fixed-dimensional vector representation $\mathcal{T} = \varepsilon(\mathbf{t}) \in \mathbb{R}^{C_{\mathcal{T}}}$ where $\varepsilon(\cdot)$ denotes a byte-pair encoding (BPE) \cite{heinzerling2017bpemb}. However, because the text embedding vector $\mathcal{T}$ differs in channel dimension from the $i$-th stage feature map $\tilde{\mathbf{X}}_{i} = \textbf{Transformer}(\textbf{C2D}_{1 \times 1} ([\textbf{Up}_{2}(\mathbf{D}_{i}), \hat{\mathbf{f}}_{i}])) \in \mathbb{R}^{C^{d}_{i} \times \frac{H}{2^{i}} \times \frac{W}{2^{i}}}$, we pass $\mathcal{T}$ through a linear embedding block—consisting of two $1 \times 1$ convolutions with a ReLU activation in between—and perform channel-wise attention as follows: \vspace{-0.15cm}
\begin{equation}
    \mathbf{D}_{i + 1} = \tilde{\mathbf{X}}_{i} \times \sigma(\textbf{C2D}_{1 \times 1}(\textbf{ReLU}(\textbf{C2D}_{1 \times 1}(\mathcal{T}))))
\end{equation}

\noindent Finally, $\mathbf{D}_{i + 1}$ is passed on to the subsequent Decoder Block, completing the processing for $i$-th decoder stage.

\begin{algorithm}[t]
\caption{Edge-Aware Difficulty Calculator}
\label{alg_DLC}
\textbf{Input}: Global prior map $\textbf{G}$ \\
\textbf{Output}: Difficulty level
\begin{algorithmic}[1] 
\STATE $\textbf{G}_{x}, \textbf{G}_{y} \leftarrow \textbf{Sobel}(\textbf{G})$
\STATE $\textbf{G}_{xx}, \textbf{G}_{xy} \leftarrow \textbf{Sobel}(\textbf{G}_{x})$
\STATE $\textbf{G}_{yx}, \textbf{G}_{yy} \leftarrow \textbf{Sobel}(\textbf{G}_{y})$
\STATE $\kappa \leftarrow (\textbf{G}_{x}^{2}\textbf{G}_{yy} - 2\textbf{G}_{x}\textbf{G}_{y} + \textbf{G}_{y}^{2}\textbf{G}_{xx}) / (\textbf{G}_{x}^{2} + \textbf{G}_{y}^{2})^{1.5}$
\STATE $\textbf{E} \leftarrow \sqrt{\textbf{G}_{x}^{2} + \textbf{G}_{y}^{2}}$
\STATE $\textbf{S} \leftarrow \sigma(\sum (\kappa \otimes \textbf{E}) / \sum \textbf{E})$
\IF{$\textbf{S} \ge 0.5$}
\STATE \textbf{return} "hard"
\ELSE
\STATE \textbf{return} "easy"
\ENDIF
\end{algorithmic}
\end{algorithm}

\noindent \textbf{Network Training.} We obtain the final forged prediction mask $\mathbf{R}_{p}$ by applying a $1 \times 1$ following sigmoid function and upsampling at the last decoder stage, and compute the overall loss as the sum of two BCE terms—one for the global prior map and another for the final prediction mask as $\mathcal{L}_{\text{total}} = \mathcal{L}_{\text{BCE}} (\mathbf{R}_{t}, \mathbf{R}_{p}) + \mathcal{L}_{\text{BCE}} (\mathbf{R}_{t}, \mathbf{Up}_{32} (\mathbf{G}))$ where $\mathbf{R}_{t}$ denotes the ground truths of the forged region, respectively. For the training phase, the parameters of M2SFormer are updated in end-to-end manner using $\mathcal{L}_{\text{total}}$. \vspace{-0.15cm}

\begin{table*}[t]
    \centering
    \scriptsize
    \setlength\tabcolsep{2.0pt} 
    \renewcommand{\arraystretch}{0.95} 
    \begin{tabular}{c|cc||cc|cc|cc|cc|cc|cc|cc}
    \hline
    \multicolumn{1}{c|}{\multirow{3}{*}{Method}} & \multicolumn{2}{c||}{Seen Domain} & \multicolumn{14}{c}{Unseen Domain} \\ \cline{2-17}
    & \multicolumn{2}{c||}{CASIAv2 \cite{pham2019hybrid}} & \multicolumn{2}{c|}{DIS25k \cite{tahir2024deep}} & \multicolumn{2}{c|}{CASIAv1 \cite{Dong2013}} & \multicolumn{2}{c|}{Columbia \cite{hsu06crfcheck}} & \multicolumn{2}{c|}{IMD2020 \cite{Novozamsky_2020_WACV}} & \multicolumn{2}{c|}{CoMoFoD \cite{tralic2013comofod}} & \multicolumn{2}{c|}{In the Wild \cite{huh2018fighting}} & \multicolumn{2}{c}{MISD \cite{kadam2021multiple}} \\ \cline{2-17}
     & DSC & mIoU & DSC & mIoU & DSC & mIoU & DSC & mIoU & DSC & mIoU & DSC & mIoU & DSC & mIoU & DSC & mIoU \\ 
     \hline
     UNet \cite{ronneberger2015u}           & 32.3 \tiny{(10.9)} & 25.8 \tiny{(9.2)} & 8.7 \tiny{(1.2)} & 5.5 \tiny{(0.8)} & 25.0 \tiny{(1.9)} & 19.5 \tiny{(1.8)} & 19.8 \tiny{(2.5)} & 12.2 \tiny{(1.7)} & 14.9 \tiny{(1.0)} & 9.7 \tiny{(0.6)} & 12.2 \tiny{(1.0)} & 7.6 \tiny{(0.7)} & 18.6 \tiny{(1.4)} & 11.9 \tiny{(1.0)} & 47.3 \tiny{(2.5)} & 34.8 \tiny{(2.2)} \\
     SegNet \cite{badrinarayanan2017segnet} & 8.1 \tiny{(6.6)} & 6.1 \tiny{(5.0)} & 1.1 \tiny{(0.7)} & 0.7 \tiny{(0.4)} & 5.0 \tiny{(1.4)} & 3.7 \tiny{(1.0)} & 7.0 \tiny{(3.7)} & 4.3 \tiny{(2.5)} & 6.3 \tiny{(2.4)} & 4.4 \tiny{(1.6)} & 3.5 \tiny{(2.0)} & 2.3 \tiny{(1.5)} & 5.8 \tiny{(2.8)} & 3.9 \tiny{(1.7)} & 20.4 \tiny{(5.7)} & 13.8 \tiny{(4.2)} \\
     MantraNet \cite{wu2019mantra} & 18.8 \tiny{(8.0)} & 11.9 \tiny{(5.7)} & 12.7 \tiny{(0.8)} & 7.4 \tiny{(0.6)} & 19.8 \tiny{(1.0)} & 12.1 \tiny{(0.8)} & 25.0 \tiny{(1.8)} & 14.9 \tiny{(1.2)} & 14.1 \tiny{(0.5)} & 8.2 \tiny{(0.3)} & 11.2 \tiny{(0.6)} & 6.5 \tiny{(0.4)} & 18.2 \tiny{(1.1)} & 10.6 \tiny{(0.8)} & 30.7 \tiny{(3.5)} & 19.2 \tiny{(2.5)} \\
     RRUNet \cite{bi2019rru} & 21.8 \tiny{(10.4)} & 15.8 \tiny{(8.4)} & 10.8 \tiny{(2.2)} & 6.9 \tiny{(1.5)} & 26.7 \tiny{(2.4)} & 18.9 \tiny{(1.7)} & 21.3 \tiny{(7.2)} & 13.9 \tiny{(5.3)} & 14.5 \tiny{(1.9)} & 9.4 \tiny{(3.3)} & 13.7 \tiny{(2.2)} & 9.0 \tiny{(1.7)} & 18.3 \tiny{(4.2)} & 11.7 \tiny{(3.0)} & 32.8 \tiny{(3.4)} & 21.4 \tiny{(2.5)} \\
     MT-SENet \cite{zhang2021multi} & 19.9 \tiny{(9.9)} & 14.6 \tiny{(7.6)} & 7.8 \tiny{(1.0)} & 4.8 \tiny{(0.7)} & 18.4 \tiny{(1.6)} & 12.8 \tiny{(1.3)} & 9.4 \tiny{(1.1)} & 5.3 \tiny{(0.6)} & 11.4 \tiny{(1.3)} & 7.2 \tiny{(1.0)} & 10.6 \tiny{(2.1)} & 6.5 \tiny{(1.5)} & 10.6 \tiny{(2.1)} & 6.5 \tiny{(1.5)} & 22.3 \tiny{(2.5)} & 14.0 \tiny{(1.8)} \\
     TransForensic \cite{hao2021transforensics} & 40.1 \tiny{(15.7)} & 32.0 \tiny{(13.9)} & 32.3 \tiny{(2.9)} & 24.1 \tiny{(2.5)} & 44.2 \tiny{(1.6)} & 35.0 \tiny{(1.7)} & \textcolor{blue}{\textbf{\textit{35.9}}} \tiny{(5.7)} & \textcolor{blue}{\textbf{\textit{25.0}}} \tiny{(4.6)} & 27.2 \tiny{(2.1)} & 19.1 \tiny{(1.6)} & 21.7 \tiny{(2.0)} & 14.3 \tiny{(1.5)} & \textcolor{blue}{\textbf{\textit{31.9}}} \tiny{(3.8)} & 22.4 \tiny{(3.0)} & 60.0 \tiny{(1.8)} & 46.5 \tiny{(2.1)} \\
     MVSSNet \cite{dong2022mvss} & 31.2 \tiny{(13.4)} & 23.4 \tiny{(11.1)} & 24.9 \tiny{(3.6)} & 17.4 \tiny{(2.9)} & 36.6 \tiny{(1.6)} & 27.3 \tiny{(1.5)} & 33.8 \tiny{(3.7)} & 23.3 \tiny{(3.0)} & 22.8 \tiny{(2.4)} & 15.2 \tiny{(1.8)} & 17.2 \tiny{(1.2)} & 10.9 \tiny{(0.9)} & 27.0 \tiny{(3.1)} & 18.3 \tiny{(2.2)} & 53.9 \tiny{(3.6)} & 17.4 \tiny{(2.9)} \\
     FBINet \cite{gu2022fbi} & 35.3 \tiny{(14.3)} & 29.2 \tiny{(12.8)} & 26.3 \tiny{(2.3)} & 20.3 \tiny{(1.9)} & 37.5 \tiny{(2.1)} & 30.8 \tiny{(1.9)} & 18.2 \tiny{(2.7)} & 11.7 \tiny{(2.1)} & 24.2 \tiny{(1.6)} & 17.5 \tiny{(1.3)} & 22.3 \tiny{(0.8)} & 15.2 \tiny{(0.5)} & 25.0 \tiny{(2.7)} & 17.8 \tiny{(2.1)} & 48.0 \tiny{(2.2)} & 35.1 \tiny{(2.0)} \\
     SegNeXt \cite{guo2022segnext} & 12.8 \tiny{(4.2)} & 8.6 \tiny{(3.0)} & 9.6 \tiny{(1.1)} & 5.6 \tiny{(0.7)} & 14.8 \tiny{(2.0)} & 9.4 \tiny{(1.5)} & 12.8 \tiny{(1.4)} & 7.3 \tiny{(0.9)} & 11.2 \tiny{(1.3)} & 6.6 \tiny{(0.8)} & 8.4 \tiny{(1.1)} & 4.8 \tiny{(0.6)} & 13.9 \tiny{(2.3)} & 8.1 \tiny{(1.5)} & 21.1 \tiny{(3.0)} & 12.7 \tiny{(2.0)} \\
     CFLNet \cite{niloy2023cfl} & 40.4 \tiny{(15.8)} & 33.4 \tiny{(14.2)} & 25.5 \tiny{(1.7)} & 18.9 \tiny{(1.3)} & 38.0 \tiny{(0.8)} & 31.1 \tiny{(0.5)} & 16.7 \tiny{(2.8)} & 10.6 \tiny{(2.0)} & 22.7 \tiny{(1.4)} & 15.8 \tiny{(1.0)} & 20.5 \tiny{(2.0)} & 14.0 \tiny{(1.5)} & 25.0 \tiny{(2.3)} & 17.3 \tiny{(1.7)} & \textcolor{blue}{\textbf{\textit{61.6}}} \tiny{(1.5)} & \textcolor{blue}{\textbf{\textit{48.3}}} \tiny{(1.6)} \\
     EITLNet \cite{guo2024effective} & 54.0 \tiny{(14.7)} & 47.9 \tiny{(14.7)} & 30.8 \tiny{(2.8)} & 25.6 \tiny{(2.6)} & \textcolor{blue}{\textbf{\textit{52.9}}} \tiny{(1.7)} & \textcolor{blue}{\textbf{\textit{46.5}}} \tiny{(1.5)} & 28.0 \tiny{(4.6)} & 20.9 \tiny{(4.0)} & 25.3 \tiny{(2.6)} & 19.7 \tiny{(2.2)} & 18.1 \tiny{(1.8)} & 12.4 \tiny{(1.4)} & 24.3 \tiny{(3.6)} & 19.0 \tiny{(3.1)} & 58.8 \tiny{(1.8)} & 45.9 \tiny{(1.8)} \\
     PIMNet \cite{bai2025pim} & \textcolor{blue}{\textbf{\textit{55.8}}} \tiny{(15.1)} & \textcolor{blue}{\textbf{\textit{48.5}}} \tiny{(14.6)} & \textcolor{blue}{\textbf{\textit{37.5}}} \tiny{(2.4)} & \textcolor{blue}{\textbf{\textit{30.1}}} \tiny{(2.3)} & 49.7 \tiny{(0.8)} & 42.2 \tiny{(1.0)} & 32.5 \tiny{(5.2)} & 23.1 \tiny{(4.3)} & \textcolor{blue}{\textbf{\textit{29.6}}} \tiny{(2.7)} & \textcolor{blue}{\textbf{\textit{22.2}}} \tiny{(2.3)} & \textcolor{blue}{\textbf{\textit{24.7}}} \tiny{(1.6)} & \textcolor{blue}{\textbf{\textit{16.8}}} \tiny{(1.4)} & 31.2 \tiny{(2.5)} & \textcolor{blue}{\textbf{\textit{22.9}}} \tiny{(2.2)} & 61.1 \tiny{(0.9)} & 48.2 \tiny{(0.9)} \\
     \hline
     \textbf{M2SFormer} & \textcolor{red}{\textbf{\underline{58.8}}} \tiny{(12.8)} & \textcolor{red}{\textbf{\underline{50.8}}} \tiny{(12.8)} & \textcolor{red}{\textbf{\underline{38.5}}} \tiny{(2.4)} & \textcolor{red}{\textbf{\underline{31.3}}} \tiny{(2.3)} & \textcolor{red}{\textbf{\underline{58.4}}} \tiny{(0.7)} & \textcolor{red}{\textbf{\underline{50.1}}} \tiny{(0.6)} & \textcolor{red}{\textbf{\underline{42.4}}} \tiny{(5.8)} & \textcolor{red}{\textbf{\underline{32.4}}} \tiny{(5.3)} & \textcolor{red}{\textbf{\underline{32.6}}} \tiny{(2.2)} & \textcolor{red}{\textbf{\underline{24.9}}} \tiny{(1.9)} & \textcolor{red}{\textbf{\underline{24.9}}} \tiny{(1.3)} & \textcolor{red}{\textbf{\underline{16.8}}} \tiny{(1.0)} & \textcolor{red}{\textbf{\underline{35.0}}} \tiny{(1.8)} & \textcolor{red}{\textbf{\underline{27.4}}} \tiny{(1.6)} & \textcolor{red}{\textbf{\underline{69.1}}} \tiny{(0.7)} & \textcolor{red}{\textbf{\underline{56.9}}} \tiny{(0.8)} \\ \cline{2-17}
     \hline
    \end{tabular} \vspace{-0.25cm}
    \caption{Segmentation results on CASIAv2 \cite{pham2019hybrid} training scheme. $(\cdot)$ denotes the standard deviations of five-fold cross-validation experiment results.} \vspace{-0.25cm}
    \label{tab:casiav2_training_scheme}
\end{table*}

\begin{table*}[t]
    \centering
    \scriptsize
    \setlength\tabcolsep{2.0pt} 
    \renewcommand{\arraystretch}{0.95} 
    \begin{tabular}{c|cc||cc|cc|cc|cc|cc|cc|cc}
    \hline
    \multicolumn{1}{c|}{\multirow{3}{*}{Method}} & \multicolumn{2}{c||}{Seen Domain} & \multicolumn{14}{c}{Unseen Domain} \\ \cline{2-17}
     & \multicolumn{2}{c||}{DIS25k \cite{tahir2024deep}} & \multicolumn{2}{c|}{CASIAv2 \cite{pham2019hybrid}} & \multicolumn{2}{c|}{CASIAv1 \cite{Dong2013}} & \multicolumn{2}{c|}{Columbia \cite{hsu06crfcheck}} & \multicolumn{2}{c|}{IMD2020 \cite{Novozamsky_2020_WACV}} & \multicolumn{2}{c|}{CoMoFoD \cite{tralic2013comofod}} & \multicolumn{2}{c|}{In the Wild \cite{huh2018fighting}} & \multicolumn{2}{c}{MISD \cite{kadam2021multiple}} \\ \cline{2-17}
     & DSC & mIoU & DSC & mIoU & DSC & mIoU & DSC & mIoU & DSC & mIoU & DSC & mIoU & DSC & mIoU & DSC & mIoU \\ 
     \hline
     UNet \cite{ronneberger2015u} & 80.9 \tiny{(13.2)} & 74.2 \tiny{(14.1)} & 15.1 \tiny{(6.4)} & 9.7 \tiny{(4.6)} & 24.0 \tiny{(0.4)} & 15.7 \tiny{(0.3)} & 28.6 \tiny{(1.1)} & 18.5 \tiny{(0.8)} & 23.1 \tiny{(0.1)} & 15.0 \tiny{(0.1)} & 13.4 \tiny{(0.7)} & 7.9 \tiny{(0.5)} & 33.9 \tiny{(0.8)} & 22.4 \tiny{(0.5)} & 40.1 \tiny{(1.3)} & 27.0 \tiny{(1.1)} \\
     SegNet \cite{badrinarayanan2017segnet} & 41.7 \tiny{(8.6)} & 32.5 \tiny{(7.4)} & 6.0 \tiny{(3.1)} & 4.1 \tiny{(2.3)} & 5.6 \tiny{(0.8)} & 3.5 \tiny{(0.5)} & 2.5 \tiny{(0.5)} & 1.3 \tiny{(0.3)} & 8.3 \tiny{(0.6)} & 5.2 \tiny{(0.4)} & 9.9 \tiny{(1.2)} & 6.6 \tiny{(0.8)} & 5.2 \tiny{(0.4)} & 3.1 \tiny{(0.3)} & 13.5 \tiny{(1.0)} & 8.2 \tiny{(0.6)} \\
     MantraNet \cite{wu2019mantra} & 69.6 \tiny{(12.1)} & 59.1 \tiny{(11.8)} & 12.7 \tiny{(4.7)} & 7.6 \tiny{(3.1)} & 18.1 \tiny{(0.4)} & 11.0 \tiny{(0.3)} & 26.5 \tiny{(3.1)} & 16.3 \tiny{(2.2)} & 18.6 \tiny{(0.6)} & 11.3 \tiny{(0.4)} & 11.9 \tiny{(0.6)} & 6.8 \tiny{(0.3)} & 26.2 \tiny{(1.3)} & 16.0 \tiny{(0.9)} & 31.9 \tiny{(1.1)} & 20.1 \tiny{(0.9)} \\
     RRUNet \cite{bi2019rru} & 76.8 \tiny{(11.3)} & 68.9 \tiny{(11.5)} & 13.4 \tiny{(6.6)} & 9.3 \tiny{(5.1)} & 21.4 \tiny{(0.9)} & 15.0 \tiny{(0.7)} & 29.9 \tiny{(2.0)} & 20.7 \tiny{(1.6)} & 17.8 \tiny{(0.7)} & 12.0 \tiny{(0.5)} & 9.4 \tiny{(2.1)} & 5.8 \tiny{(1.5)} & 23.7 \tiny{(1.2)} & 16.0 \tiny{(0.9)} & 32.2 \tiny{(1.6)} & 21.1 \tiny{(1.2)} \\
     MT-SENet \cite{zhang2021multi} & 77.8 \tiny{(12.7)} & 70.4 \tiny{(13.4)} & 12.8 \tiny{(5.7)} & 8.2 \tiny{(4.1)} & 22.0 \tiny{(0.7)} & 14.2 \tiny{(0.5)} & 26.8 \tiny{(1.6)} & 17.2 \tiny{(1.2)} & 1.9 \tiny{(0.2)} & 12.4 \tiny{(0.2)} & 13.1 \tiny{(1.0)} & 7.7 \tiny{(0.6)} & 26.5 \tiny{(0.9)} & 16.9 \tiny{(0.6)} & 34.2 \tiny{(1.2)} & 22.3 \tiny{(0.5)} \\
     TransForensic \cite{hao2021transforensics} & 83.4 \tiny{(6.9)} & 76.2 \tiny{(7.6)} & 21.2 \tiny{(11.1)} & 16.4 \tiny{(9.7)} & 30.4 \tiny{(0.7)} & 23.9 \tiny{(0.5)} & 33.0 \tiny{(1.7)} & 23.4 \tiny{(1.3)} & 27.0 \tiny{(0.6)} & 19.8 \tiny{(0.5)} & 18.1 \tiny{(2.2)} & 12.1 \tiny{(1.8)} & 31.0 \tiny{(0.8)} & 22.3 \tiny{(0.7)} & 45.7 \tiny{(1.4)} & 33.2 \tiny{(1.2)} \\
     MVSSNet \cite{dong2022mvss} & 70.0 \tiny{(4.3)} & 60.1 \tiny{(4.2)} & 19.8 \tiny{(10.2)} & 14.4 \tiny{(8.1)} & 30.0 \tiny{(1.3)} & 22.6 \tiny{(1.0)} & 35.7 \tiny{(4.7)} & 26.4 \tiny{(3.8)} & 24.7 \tiny{(1.3)} & 17.4 \tiny{(0.9)} & 17.5 \tiny{(1.5)} & 11.3 \tiny{(1.0)} & 30.7 \tiny{(2.9)} & 22.0 \tiny{(2.0)} & \textcolor{red}{\textbf{\underline{47.7}}} \tiny{(2.1)} & \textcolor{red}{\textbf{\underline{34.8}}} \tiny{(1.8)} \\
     FBINet \cite{gu2022fbi} & 81.6 \tiny{(5.4)} & 74.7 \tiny{(5.8)} & 17.9 \tiny{(10.3)} & 13.8 \tiny{(8.8)} & 21.0 \tiny{(0.7)} & 15.9 \tiny{(0.6)} & 13.9 \tiny{(1.7)} & 8.7 \tiny{(1.3)} & 21.6 \tiny{(0.5)} & 15.4 \tiny{(0.5)} & 15.0 \tiny{(1.7)} & 9.9 \tiny{(1.3)} & 17.9 \tiny{(1.5)} & 12.3 \tiny{(1.2)} & 32.4 \tiny{(1.7)} & 21.4 \tiny{(1.4)} \\
     SegNeXt \cite{guo2022segnext} & 72.9 \tiny{(11.0)} & 63.7 \tiny{(11.2)} & 13.4 \tiny{(5.7)} & 8.4 \tiny{(4.0)} & 23.7 \tiny{(0.3)} & 15.3 \tiny{(0.2)} & 35.5 \tiny{(1.1)} & 24.1 \tiny{(0.8)} & 21.5 \tiny{(0.2)} & 13.8 \tiny{(0.1)} & 13.6 \tiny{(1.0)} & 8.1 \tiny{(0.6)} & \textcolor{blue}{\textbf{\textit{32.2}}} \tiny{(0.5)} & 21.2 \tiny{(0.4)} & 41.5 \tiny{(0.9)} & 28.1 \tiny{(0.8)} \\
     CFLNet \cite{niloy2023cfl} & 79.8 \tiny{(6.7)} & 71.5 \tiny{(7.3)} & 20.2 \tiny{(10.1)} & 15.0 \tiny{(8.3)} & 29.0 \tiny{(0.6)} & 22.2 \tiny{(0.5)} & 30.0 \tiny{(3.0)} & 20.9 \tiny{(2.6)} & 26.1 \tiny{(0.4)} & 18.5 \tiny{(0.4)} & 17.9 \tiny{(0.9)} & 11.6 \tiny{(0.8)} & 29.8 \tiny{(1.1)} & 20.9 \tiny{(0.8)} & \textcolor{blue}{\textbf{\textit{46.0}}} \tiny{(1.5)} & \textcolor{blue}{\textbf{\textit{33.4}}} \tiny{(1.3)} \\
     EITLNet \cite{guo2024effective} & \textcolor{red}{\textbf{\underline{90.6}}} \tiny{(3.9)} & \textcolor{red}{\textbf{\underline{85.7}}} \tiny{(4.1)} & 25.4 \tiny{(13.7)} & \textcolor{blue}{\textbf{\textit{21.3}}} \tiny{(13.1)} & \textcolor{blue}{\textbf{\textit{36.2}}} \tiny{(0.3)} & \textcolor{blue}{\textbf{\textit{31.1}}} \tiny{(0.3)} & 31.6 \tiny{(2.4)} & 24.0 \tiny{(2.2)} & \textcolor{blue}{\textbf{\textit{30.0}}} \tiny{(0.4)} & \textcolor{blue}{\textbf{\textit{23.8}}} \tiny{(0.2)} & 19.9 \tiny{(2.8)} & 14.2 \tiny{(2.1)} & 29.2 \tiny{(0.8)} & 23.1 \tiny{(0.6)} & 39.0 \tiny{(0.8)} & 27.2 \tiny{(0.7)} \\
     PIMNet \cite{bai2025pim} & \textcolor{blue}{\textbf{\textit{88.8}}} \tiny{(4.7)} & \textcolor{blue}{\textbf{\textit{83.3}}} \tiny{(5.0)} & \textcolor{blue}{\textbf{\textit{25.4}}} \tiny{(12.2)} & 20.6 \tiny{(11.0)} & 34.5 \tiny{(1.2)} & 28.1 \tiny{(0.9)} & \textcolor{red}{\textbf{\underline{38.6}}} \tiny{(5.4)} & \textcolor{red}{\textbf{\underline{29.6}}} \tiny{(4.7)} & 29.7 \tiny{(0.9)} & 22.7 \tiny{(0.8)} & \textcolor{blue}{\textbf{\textit{21.6}}} \tiny{(0.2)} & \textcolor{blue}{\textbf{\textit{14.9}}} \tiny{(0.5)} & \textcolor{red}{\textbf{\underline{33.4}}} \tiny{(2.4)} & \textcolor{red}{\textbf{\underline{25.0}}} \tiny{(1.9)} & 42.5 \tiny{(2.1)} & 30.2 \tiny{(1.8)} \\
     \hline
     \textbf{M2SFormer} & 87.7 \tiny{(3.2)} & 81.5 \tiny{(3.2)} & \textcolor{red}{\textbf{\underline{26.7}}} \tiny{(14.3)} & \textcolor{red}{\textbf{\underline{22.3}}} \tiny{(13.6)} & \textcolor{red}{\textbf{\underline{40.1}}} \tiny{(1.3)} & \textcolor{red}{\textbf{\underline{34.5}}} \tiny{(1.1)} & \textcolor{blue}{\textbf{\textit{37.3}}} \tiny{(2.5)} & \textcolor{blue}{\textbf{\textit{28.8}}} \tiny{(2.0)} & \textcolor{red}{\textbf{\underline{31.9}}} \tiny{(0.8)} & \textcolor{red}{\textbf{\underline{25.2}}} \tiny{(0.6)} & \textcolor{red}{\textbf{\underline{31.9}}} \tiny{(0.6)} & \textcolor{red}{\textbf{\underline{25.2}}} \tiny{(0.4)} & 31.7 \tiny{(1.1)} & \textcolor{blue}{\textbf{\textit{24.4}}} \tiny{(0.9)} & 43.2 \tiny{(1.6)} & 30.7 \tiny{(1.5)} \\ \cline{2-17}
     \hline
    \end{tabular} \vspace{-0.25cm}
    \caption{Segmentation results on DIS25k \cite{tahir2024deep} training scheme. $(\cdot)$ denotes the standard deviations of five-fold cross-validation experiment results.} \vspace{-0.75cm}
    \label{tab:dis25k_training_scheme}
\end{table*}

\section{Experiment Results} \vspace{-0.15cm}

\subsection{Experiment Settings} \vspace{-0.15cm}
To evaluate the generalization capabilities of each model, we adopt two training schemes—CASIAv2 and DIS25k—and subsequently assess cross-domain performance on six external datasets: CASIAv1, Columbia, IMD2020, CoMoFoD, In the Wild, and MISD. We want to clarify that these datasets are not used for training. Due to the page limit, we present the detailed dataset description (Tab. \ref{tab:dataset_summary}) and split information for each training scheme in the Appendix (Section \ref{appendix_dataset_descriptions}). To evaluate the performance of each model, we selected three metrics, the Dice Score Coefficient (DSC) \cite{milletari2016v}, and mean Intersection over Union (mIoU), which are widely used in the field of image segmentation. We present the detailed metric description in the Appendix (Section \ref{appendix_metric_descriptions}) due to the page limits.

\begin{figure}[t]
    \centering
    \includegraphics[width=0.48\textwidth]{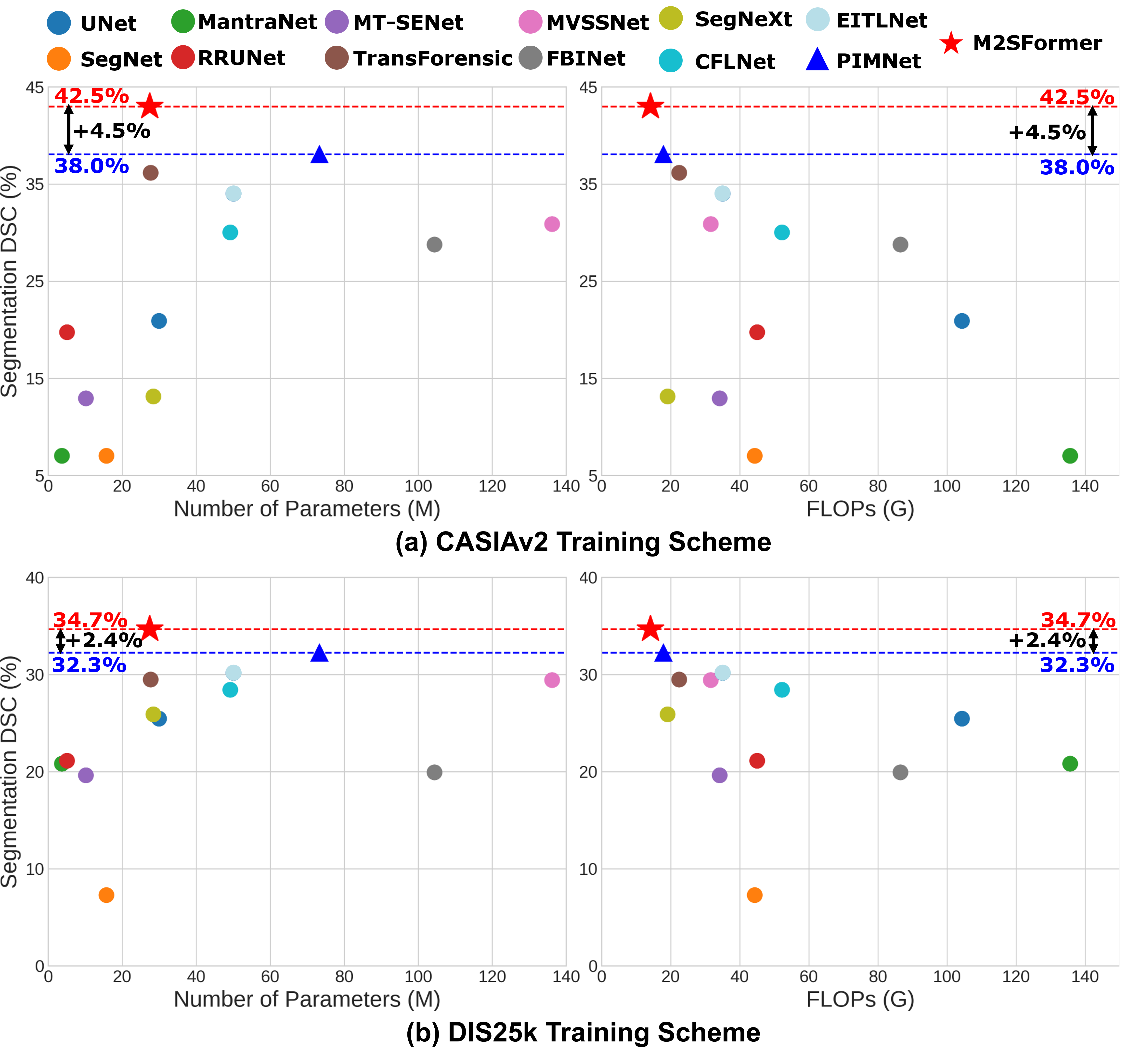} \vspace{-0.75cm}
        \caption{Comparison of parameters (M), and FLOPs (G) vs segmentation performance (DSC) on the average for the (a) CASIAv2 training scheme and (b) DIS25k training scheme. The efficiency metrics are measured at resolution $256 \times 256$.} \vspace{-0.5cm}
    \label{fig:EfficiencyAnalysis}
\end{figure}

\begin{figure*}[t]
    \centering
    \includegraphics[width=\textwidth]{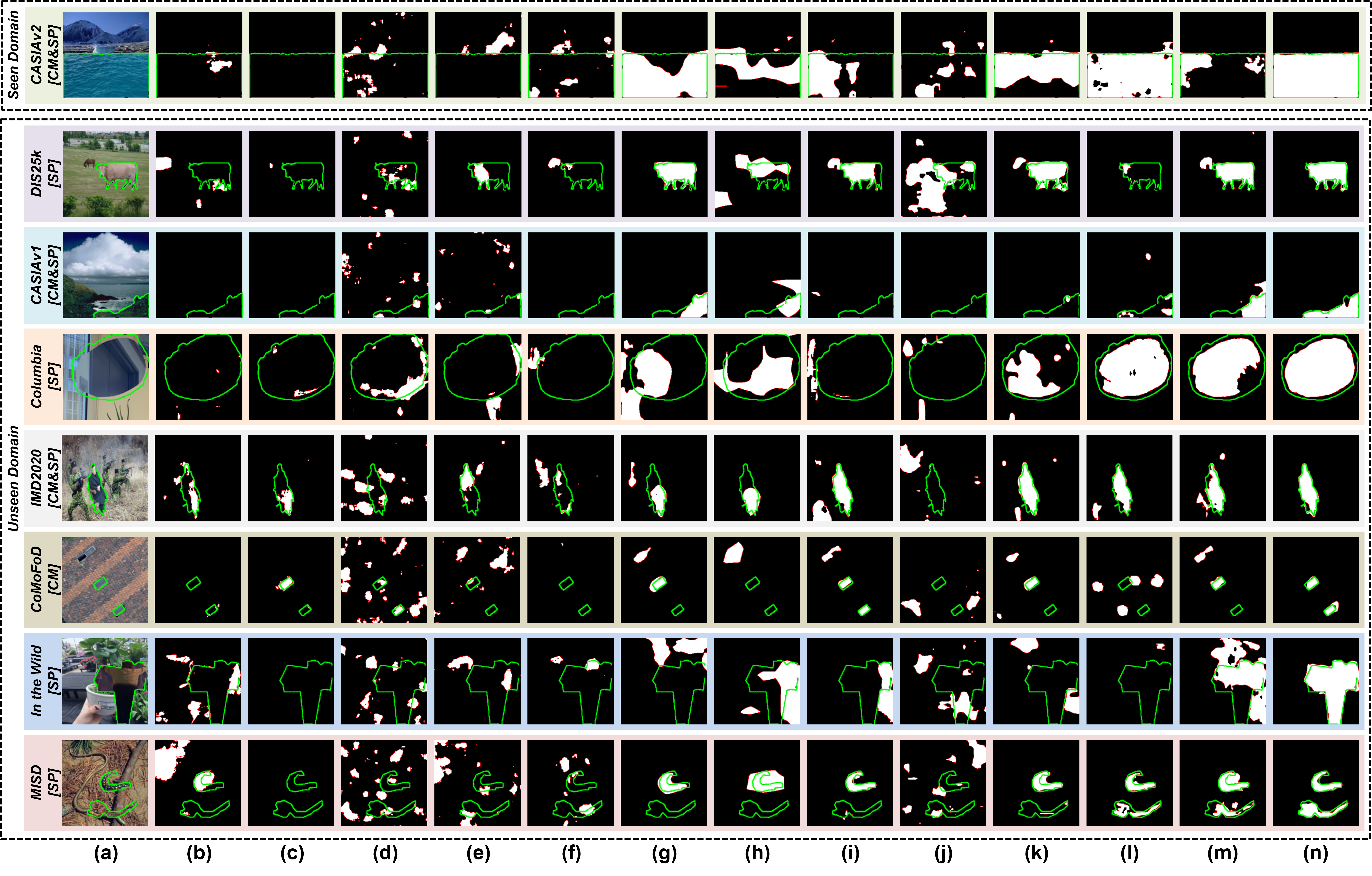} \vspace{-0.75cm}
        \caption{Qualitative comparison of other methods and M2SFormer with CASIAv2 training scheme.  (a) Input images with ground truth. (b) UNet \cite{ronneberger2015u}. (c) SegNet \cite{badrinarayanan2017segnet}. (d) MantraNet \cite{wu2019mantra}. (e) RRUNet \cite{bi2019rru}. (f) MT-SENet \cite{zhang2021multi}. (g) TransForensic \cite{hao2021transforensics}. (h) MVSSNet \cite{dong2022mvss}. (i) FBINet \cite{gu2022fbi}. (j) SegNeXt \cite{guo2022segnext}. (k) CFLNet \cite{niloy2023cfl}. (l) EITLNet \cite{guo2024effective}. (m) PIMNet \cite{bai2025pim}. (n) \textbf{M2SFormer (Ours)}. \textcolor{green}{\textbf{Green}} and \textcolor{red}{\textbf{Red}} lines denote the boundaries of the ground truth and prediction, respectively. And, the “SP” and “CM” in square brackets represent the types of forgery in the dataset: “SP” stands for Splicing, while “CM” denotes Copy-Move. We also provide the qualtitative results on DIS25k in the Appendix (Fig. \ref{fig:SupQualitativeResults})}
    \label{fig:QualitativeResults} \vspace{-0.5cm}
\end{figure*}

We compared the proposed \textbf{M2SFormer (Ours)} with fifteen representative forgery localization models, including UNet  \cite{ronneberger2015u}, SegNet \cite{badrinarayanan2017segnet}, MantraNet \cite{wu2019mantra}, RRUNet \cite{bi2019rru}, MT-SENet \cite{zhang2021multi}, TransForensic \cite{hao2021transforensics}, MVSSNet \cite{dong2022mvss}, FBINet \cite{gu2022fbi}, SegNeXt \cite{guo2022segnext}, CFLNet \cite{niloy2023cfl}, EITLNet \cite{guo2024effective}, and PIMNet \cite{bai2025pim}. Because no standardized training and evaluation datasets exist for measuring the generalization ability of forgery localization methods \cite{guo2024effective, bai2025pim}, we trained all models using the same training set and evaluated them on an identical test set to ensure a fair comparison. In all results, we report the mean performance of five-fold cross-validation results for reliability. In all tables, \textcolor{red}{\textbf{\underline{Red}}} and \textcolor{blue}{\textbf{\textit{Blue}}} are the first and second best performance results, respectively.

We started with an initial learning rate of $10^{-4}$ using the Adam \cite{kingma2014adam} optimizer and reduced the parameters of each model to $10^{-6}$ using a cosine annealing learning rate scheduler \cite{loshchilov2016sgdr}. We optimized each model using a batch size of 32 and trained them for 100 epochs. Because images in each dataset have different resolutions, all images were resized to $256 \times 256$. Additionally, we would like to clarify that we used identical settings to train each model.
 
\noindent\textbf{Hyperparameters of M2SFormer.} We use the PVT-v2-B2 ($C^{e} = \{ 64, 128, 320, 512 \}$ and $C^{d} = \{ 256, 128, 64 \}$) as the backbone of encoder and decoder. Key hyperparameters for M2SFormer on all datasets were set to $C_{r} = 64$ for efficiency and $(H_{t}, W_{t}) = (\frac{H}{8}, \frac{W}{8}), F = 16, L = 3$ in M2S attention block, and $C_{\mathcal{T}} = 300, \tau = 0.5$ in Edge-Aware DGA-based Transformer Decoder.  \vspace{-0.15cm}

\subsection{Comparison with State-of-the-art models}
\noindent \textbf{Quantitative Results Analysis.} As shown in Tab. \ref{tab:casiav2_training_scheme} and Tab. \ref{tab:dis25k_training_scheme}, our method achieves robust performance in unseen domains under both training schemes, demonstrating its effectiveness across varying experimental conditions. Fig. \ref{fig:EfficiencyAnalysis} indicates that M2SFormer contains only 14.0G FLOPs, which has apparent advantages regarding computational efficiency. Due to the page limit, we provide detailed number of parameters and FLOPs for each model in Appendix (Tab. \ref{tab:efficiency_analysis}).  When directly applied to forgery localization, popular segmentation models like UNet, SegNet, and SegNeXt—though widely used in other domains—exhibit notably low performance. When compared with RRUNet and MT-SENet—both representative UNet-like encoder–decoder architectures for forgery localization—M2SFormer demonstrates substantial gains on CASIAv2. Furthermore, M2SFormer outperforms FBINet, which applies a 2D DCT to input images for forged trace detection, while requiring fewer parameters and FLOPs under both training schemes. By contrast, EITLNet likewise integrates frequency-based cues and Transformers but relies on dual encoders, leading to relatively higher parameter counts and computational overhead. Moreover, its MLP-based decoding structure provides lower representational capacity compared to our DGA-based Transformer Decoder. \textit{Consequently, M2SFormer achieves higher performance with fewer parameters, underscoring the efficiency and enhanced representational power of the integration of M2S attention block and DGA-based Transformer Decoder}.

\noindent \textbf{Qualitative Results Analysis.} Fig. \ref{fig:QualitativeResults} illustrates the qualitative results of the various models on the CASIAv2 training scheme. UNet, SegNet, and SegNeXt struggle to capture subtle boundary cues in manipulated regions, leading to coarse or incomplete masks. CNN-based methods like MantraNet, RRUNet, and MT-SENet show improvements by leveraging learned features for pixel-level predictions, but they often leave blurred or incorrect boundaries in complex forgeries. Transformer-driven or frequency-based approaches such as TransForensic, FBINet, CFLNet, EITLNet, and PIMNet further enhance localization by modeling global context or focusing on high-frequency discrepancies, yet they still exhibit boundary gaps when domain shifts occur. Overall, the synergy of the \textit{M2S Attention Block}, which fuses spatial and frequency-domain cues, and the \textit{DGA-based Transformer Decoder}, driven by adaptive difficulty measurement, provides more precise boundary delineation and robust cross-domain generalization. \textit{As a result, \textbf{M2SFormer} outperforms both conventional segmentation-based and specialized forgery localization frameworks, demonstrating sharper and more accurate forged masks even under significant domain shifts}. \vspace{-0.15cm}

\begin{table}[t]
    \centering
    \scriptsize
    \setlength\tabcolsep{2pt} 
    \renewcommand{\arraystretch}{0.95} 
    \begin{tabular}{c|cc|cc|cc|cc|c|c}
    \hline
    Setting & \multicolumn{2}{c|}{Spectrum} & \multicolumn{2}{c|}{Scale} & \multicolumn{2}{c|}{Seen} & \multicolumn{2}{c|}{Unseen} & \multicolumn{1}{c|}{\multirow{2}{*}{Param (M)}} & \multicolumn{1}{c}{\multirow{2}{*}{FLOPs (G)}} \\ \cline{2-9}
    Name & S & M & S & M & DSC & mIoU & DSC & mIoU &  & \\
    \hline
    S0 & \cmark & & \cmark &  & \textcolor{blue}{\textbf{\textit{56.3}}} & \textcolor{blue}{\textbf{\textit{49.1}}} & 27.1 & 24.2 & 26.2MB & 13.8GB \\
    \hline
    S1 & \cmark & & & \cmark  & 55.5 & 48.6 & 33.6 & 28.1 & 27.4MB & 14.2GB \\
    S2 & & \cmark & \cmark &  & 55.9 & 49.0 & \textcolor{blue}{\textbf{\textit{36.1}}} & \textcolor{blue}{\textbf{\textit{31.3}}} & 26.2MB & 13.8GB \\
    \hline
    \textbf{S3 (Ours)} & & \cmark & & \cmark & \textcolor{red}{\textbf{\underline{58.8}}} & \textcolor{red}{\textbf{\underline{50.8}}} & \textcolor{red}{\textbf{\underline{43.0}}} & \textcolor{red}{\textbf{\underline{34.3}}} & 27.4MB & 14.2GB \\
    \hline
    \end{tabular} \vspace{-0.25cm}
    \caption{Ablation study of M2S attention block in skip connection on \textit{Seen} (CASIAv2 \cite{pham2019hybrid}) and \textit{Unseen} datasets (Other test datasets). “S” and “M” denote Single and Multi, respectively.} \vspace{-0.5cm}
    \label{tab:ablation_m2s_attention_block}
\end{table}

\subsection{Ablation Study on M2SFormer} \vspace{-0.15cm}
We conducted ablation studies on the CASIAv2 training scheme to demonstrate the effectiveness of M2S attention block and DGA-based Transformer decoder. \textit{We want to clarify that the experiment settings for the ablation study are identical to the main experiments for fair comparison}.

\noindent \textbf{Ablation Study on M2S Attention Block.} Table \ref{tab:ablation_m2s_attention_block} presents an ablation study on the M2S attention block. The results show a clear performance improvement when incorporating both multi-spectral and multi-scale attention mechanisms. S0, which includes only single-spectral and single-scale components, achieves the lowest unseen domain. S1 (incorporating multi-scale attention) improves unseen performance significantly, demonstrating the importance of multi-scale information. S2, which replaces multi-spectral with single-scale attention, further enhances generalization, proving that multi-spectral attention contributes to robustness on the unseen domain. Finally, S3 (Ours), integrating both multi-spectral and multi-scale attention, achieves the highest performance across all metrics, confirming that the full M2S attention mechanism maximizes both seen and unseen domain effectiveness. \textit{Despite a slight increase in parameters and FLOPs, the performance gain justifies the trade-off, highlighting M2SFormer’s efficiency in balancing accuracy and computational cost}. 

\noindent \textbf{Ablation Study on DGA-based Transformer Decoder.} Table \ref{tab:ablation_dga_transformer_decoder} presents an ablation study on the DGA-based Transformer across three settings: (1) No DGA, (2) Simple DC + DGA, and (3) Edge-Aware Difficulty Calculator (EADC) + DGA. The baseline without DGA performs moderately on the seen domain (DSC 55.5, mIoU 49.5), but its performance drops considerably for unseen data (DSC 32.3, mIoU 26.1). While introducing a simple difficulty calculator (Simple DC + DGA) yields slight improvements in the seen domain, it remains inadequate for handling unseen datasets. By contrast, the EADC + DGA combination achieves the highest overall performance (DSC 58.8, mIoU 50.8 on seen; DSC 43.0, mIoU 34.3 on unseen), underscoring the effectiveness of focusing on boundary transitions—often indicative of forgery artifacts—when estimating difficulty. \textit{This edge-centric approach more reliably pinpoints challenging cues, such as abrupt texture changes or altered contours, thereby enhancing the model’s ability to generalize across various types of tampering and domain shifts.} Consequently, the substantial gains observed with ECDC + DGA validate the role of edge-centric difficulty measurement in bolstering cross-domain robustness for forgery localization. \vspace{-0.15cm}
\begin{table}[t]
    \centering
    \scriptsize
    \setlength\tabcolsep{4pt} 
    \renewcommand{\arraystretch}{0.95} 
    \begin{tabular}{c|cc|cc|c|c}
    \hline
    Setting & \multicolumn{2}{c|}{Seen} & \multicolumn{2}{c|}{Unseen} & \multicolumn{1}{c|}{\multirow{2}{*}{Param (M)}} & \multicolumn{1}{c}{\multirow{2}{*}{FLOPs (G)}}\\ \cline{2-5}
    Name & DSC & mIoU & DSC & mIoU &  & \\
    \hline
    No DGA & 55.5 & 49.5 & \textcolor{blue}{\textbf{\textit{32.3}}} & \textcolor{blue}{\textbf{\textit{26.1}}} & 26.9MB & 13.0GB \\
    Simple DC + DGA & \textcolor{blue}{\textbf{\textit{56.1}}} & \textcolor{blue}{\textbf{\textit{50.2}}} & 30.8 & 24.8 & 27.4MB & 14.0GB \\
    \hline
    EADC + DGA & \textcolor{red}{\textbf{\underline{58.8}}} & \textcolor{red}{\textbf{\underline{50.8}}} & \textcolor{red}{\textbf{\underline{43.0}}} & \textcolor{red}{\textbf{\underline{34.3}}} & 27.4MB & 14.2GB \\
    \hline
    \end{tabular} \vspace{-0.25cm}
    \caption{Ablation study of DGA-based Transformer on \textit{Seen} (CASIAv2 \cite{pham2019hybrid}) and \textit{Unseen} datasets (Other test datasets). ECDC denotes Edge-Centric Difficulty Calculator.} \vspace{-0.5cm}
    \label{tab:ablation_dga_transformer_decoder}
\end{table}

\section{Conclusion} \vspace{-0.15cm}

In this paper, we tackled the challenge of generalizable forgery localization by introducing \textbf{M2SFormer}, a Transformer encoder-based framework that jointly leverages multi-spectral and multi-scale attention within the difficulty-guided attention-based Transformer Decoder. Unlike earlier methods that separately process spatial and frequency cues, M2SFormer integrates these two domains efficiently through skip connections enriched with global Transformer context, allowing for both coarse and fine-grained forgery traces to be captured. This comprehensive fusion directly contributes to the model’s adaptability in detecting diverse manipulation types, including subtle or previously unseen forgeries. Moreover, the curvature-based difficulty-guided attention effectively quantifies each test instance’s difficulty, guiding the text-based attention module to focus on intricate details that traditional decoders might overlook. Through extensive experiments on multiple benchmark datasets, M2SFormer demonstrates not only higher accuracy but also stronger cross-domain generalization than state-of-the-art methods, underscoring the effectiveness of its unified spatial-frequency attention and contextual guidance. By dynamically exploiting frequency nuances, spatial structures, and difficulty indicators, our approach significantly enhances boundary delineation and reduces the risk of missing subtle clues. Consequently, our works offers a promising direction for future research in robust, scalable image forgery localization—a field that is increasingly vital for maintaining trust and integrity in digital visual content.
\section*{Acknowledgements}

This work was supported in part by Institute of Information and communications Technology Planning \& Evaluation (IITP) grant funded by the Korea government (MSIT) (No.RS-2022-00155915, Artificial Intelligence Convergence Innovation Human Resources Development (Inha University) and in part by Inha University Research Grant.
{
    \small
    \bibliographystyle{ieeenat_fullname}
    \bibliography{main}

\begin{thebibliography}{79}
\providecommand{\natexlab}[1]{#1}
\providecommand{\url}[1]{\texttt{#1}}
\expandafter\ifx\csname urlstyle\endcsname\relax
  \providecommand{\doi}[1]{doi: #1}\else
  \providecommand{\doi}{doi: \begingroup \urlstyle{rm}\Url}\fi

\bibitem[Ahmed et~al.(1974)Ahmed, Natarajan, and Rao]{ahmed1974discrete}
Nasir Ahmed, T\_ Natarajan, and Kamisetty~R Rao.
\newblock Discrete cosine transform.
\newblock \emph{IEEE transactions on Computers}, 100\penalty0 (1):\penalty0 90--93, 1974.

\bibitem[Ahmed and Chua(2023)]{ahmed2023perception}
Saifuddin Ahmed and Hui~Wen Chua.
\newblock Perception and deception: Exploring individual responses to deepfakes across different modalities.
\newblock \emph{Heliyon}, 9\penalty0 (10), 2023.

\bibitem[Attneave(1954)]{attneave1954some}
Fred Attneave.
\newblock Some informational aspects of visual perception.
\newblock \emph{Psychological review}, 61\penalty0 (3):\penalty0 183, 1954.

\bibitem[Azad et~al.(2021)Azad, Bozorgpour, Asadi-Aghbolaghi, Merhof, and Escalera]{azad2021deep}
Reza Azad, Afshin Bozorgpour, Maryam Asadi-Aghbolaghi, Dorit Merhof, and Sergio Escalera.
\newblock Deep frequency re-calibration u-net for medical image segmentation.
\newblock In \emph{Proceedings of the IEEE/CVF International Conference on Computer Vision}, pages 3274--3283, 2021.

\bibitem[Badrinarayanan et~al.(2017)Badrinarayanan, Kendall, and Cipolla]{badrinarayanan2017segnet}
Vijay Badrinarayanan, Alex Kendall, and Roberto Cipolla.
\newblock Segnet: A deep convolutional encoder-decoder architecture for image segmentation.
\newblock \emph{IEEE transactions on pattern analysis and machine intelligence}, 39\penalty0 (12):\penalty0 2481--2495, 2017.

\bibitem[Bai et~al.(2025)Bai, Wang, Han, Hou, Wang, and Pang]{bai2025pim}
Ningning Bai, Xiaofeng Wang, Ruidong Han, Jianpeng Hou, Yihang Wang, and Shanmin Pang.
\newblock Pim-net: Progressive inconsistency mining network for image manipulation localization.
\newblock \emph{Pattern Recognition}, 159:\penalty0 111136, 2025.

\bibitem[Bi et~al.(2019)Bi, Wei, Xiao, and Li]{bi2019rru}
Xiuli Bi, Yang Wei, Bin Xiao, and Weisheng Li.
\newblock Rru-net: The ringed residual u-net for image splicing forgery detection.
\newblock In \emph{Proceedings of the IEEE/CVF Conference on Computer Vision and Pattern Recognition Workshops}, pages 0--0, 2019.

\bibitem[Chen et~al.(2021)Chen, Dong, Ji, Cao, and Li]{chen2021image}
Xinru Chen, Chengbo Dong, Jiaqi Ji, Juan Cao, and Xirong Li.
\newblock Image manipulation detection by multi-view multi-scale supervision.
\newblock In \emph{Proceedings of the IEEE/CVF international conference on computer vision}, pages 14185--14193, 2021.

\bibitem[Chen et~al.(2019)Chen, Fan, Xu, Yan, Kalantidis, Rohrbach, Yan, and Feng]{chen2019drop}
Yunpeng Chen, Haoqi Fan, Bing Xu, Zhicheng Yan, Yannis Kalantidis, Marcus Rohrbach, Shuicheng Yan, and Jiashi Feng.
\newblock Drop an octave: Reducing spatial redundancy in convolutional neural networks with octave convolution.
\newblock In \emph{Proceedings of the IEEE/CVF international conference on computer vision}, pages 3435--3444, 2019.

\bibitem[Chierchia et~al.(2014)Chierchia, Poggi, Sansone, and Verdoliva]{chierchia2014bayesian}
Giovanni Chierchia, Giovanni Poggi, Carlo Sansone, and Luisa Verdoliva.
\newblock A bayesian-mrf approach for prnu-based image forgery detection.
\newblock \emph{IEEE Transactions on Information Forensics and Security}, 9\penalty0 (4):\penalty0 554--567, 2014.

\bibitem[Dai et~al.(2021)Dai, Gieseke, Oehmcke, Wu, and Barnard]{dai2021attentional}
Yimian Dai, Fabian Gieseke, Stefan Oehmcke, Yiquan Wu, and Kobus Barnard.
\newblock Attentional feature fusion.
\newblock In \emph{Proceedings of the IEEE/CVF winter conference on applications of computer vision}, pages 3560--3569, 2021.

\bibitem[Dirik and Memon(2009)]{dirik2009image}
Ahmet~Emir Dirik and Nasir Memon.
\newblock Image tamper detection based on demosaicing artifacts.
\newblock In \emph{2009 16th IEEE International Conference on Image Processing (ICIP)}, pages 1497--1500. IEEE, 2009.

\bibitem[Dong et~al.(2022)Dong, Chen, Hu, Cao, and Li]{dong2022mvss}
Chengbo Dong, Xinru Chen, Ruohan Hu, Juan Cao, and Xirong Li.
\newblock Mvss-net: Multi-view multi-scale supervised networks for image manipulation detection.
\newblock \emph{IEEE Transactions on Pattern Analysis and Machine Intelligence}, 45\penalty0 (3):\penalty0 3539--3553, 2022.

\bibitem[Dong et~al.(2013)Dong, Wang, and Tan]{Dong2013}
Jing Dong, Wei Wang, and Tieniu Tan.
\newblock {CASIA} image tampering detection evaluation database.
\newblock In \emph{2013 {IEEE} China Summit and International Conference on Signal and Information Processing}. {IEEE}, 2013.

\bibitem[Dosovitskiy et~al.(2021)Dosovitskiy, Beyer, Kolesnikov, Weissenborn, Zhai, Unterthiner, Dehghani, Minderer, Heigold, Gelly, Uszkoreit, and Houlsby]{dosovitskiy2021an}
Alexey Dosovitskiy, Lucas Beyer, Alexander Kolesnikov, Dirk Weissenborn, Xiaohua Zhai, Thomas Unterthiner, Mostafa Dehghani, Matthias Minderer, Georg Heigold, Sylvain Gelly, Jakob Uszkoreit, and Neil Houlsby.
\newblock An image is worth 16x16 words: Transformers for image recognition at scale.
\newblock In \emph{International Conference on Learning Representations}, 2021.

\bibitem[Ferrara et~al.(2012)Ferrara, Bianchi, De~Rosa, and Piva]{ferrara2012image}
Pasquale Ferrara, Tiziano Bianchi, Alessia De~Rosa, and Alessandro Piva.
\newblock Image forgery localization via fine-grained analysis of cfa artifacts.
\newblock \emph{IEEE Transactions on Information Forensics and Security}, 7\penalty0 (5):\penalty0 1566--1577, 2012.

\bibitem[Gao et~al.(2019)Gao, Cheng, Zhao, Zhang, Yang, and Torr]{gao2019res2net}
Shang-Hua Gao, Ming-Ming Cheng, Kai Zhao, Xin-Yu Zhang, Ming-Hsuan Yang, and Philip Torr.
\newblock Res2net: A new multi-scale backbone architecture.
\newblock \emph{IEEE transactions on pattern analysis and machine intelligence}, 43\penalty0 (2):\penalty0 652--662, 2019.

\bibitem[Gu et~al.(2022)Gu, Nam, and Lee]{gu2022fbi}
A-Rom Gu, Ju-Hyeon Nam, and Sang-Chul Lee.
\newblock Fbi-net: Frequency-based image forgery localization via multitask learning with self-attention.
\newblock \emph{IEEE Access}, 10:\penalty0 62751--62762, 2022.

\bibitem[Guo et~al.(2024)Guo, Zhu, and Cao]{guo2024effective}
Kun Guo, Haochen Zhu, and Gang Cao.
\newblock Effective image tampering localization via enhanced transformer and co-attention fusion.
\newblock In \emph{ICASSP 2024-2024 IEEE International Conference on Acoustics, Speech and Signal Processing (ICASSP)}, pages 4895--4899. IEEE, 2024.

\bibitem[Guo et~al.(2022)Guo, Lu, Hou, Liu, Cheng, and Hu]{guo2022segnext}
Meng-Hao Guo, Cheng-Ze Lu, Qibin Hou, Zhengning Liu, Ming-Ming Cheng, and Shi-Min Hu.
\newblock Segnext: Rethinking convolutional attention design for semantic segmentation.
\newblock \emph{Advances in Neural Information Processing Systems}, 35:\penalty0 1140--1156, 2022.

\bibitem[Guo et~al.(2023)Guo, Liu, Ren, Grosz, Masi, and Liu]{guo2023hierarchical}
Xiao Guo, Xiaohong Liu, Zhiyuan Ren, Steven Grosz, Iacopo Masi, and Xiaoming Liu.
\newblock Hierarchical fine-grained image forgery detection and localization.
\newblock In \emph{Proceedings of the IEEE/CVF Conference on Computer Vision and Pattern Recognition}, pages 3155--3165, 2023.

\bibitem[Han et~al.(2024)Han, Wang, Bai, Wang, Hou, and Xue]{han2024hdf}
Ruidong Han, Xiaofeng Wang, Ningning Bai, Yihang Wang, Jianpeng Hou, and Jianru Xue.
\newblock Hdf-net: Capturing homogeny difference features to localize the tampered image.
\newblock \emph{IEEE Transactions on Pattern Analysis and Machine Intelligence}, 2024.

\bibitem[Hao et~al.(2021)Hao, Zhang, Yang, Xie, and Pu]{hao2021transforensics}
Jing Hao, Zhixin Zhang, Shicai Yang, Di Xie, and Shiliang Pu.
\newblock Transforensics: image forgery localization with dense self-attention.
\newblock In \emph{Proceedings of the IEEE/CVF International Conference on Computer Vision}, pages 15055--15064, 2021.

\bibitem[He et~al.(2016)He, Zhang, Ren, and Sun]{he2016deep}
Kaiming He, Xiangyu Zhang, Shaoqing Ren, and Jian Sun.
\newblock Deep residual learning for image recognition.
\newblock In \emph{Proceedings of the IEEE conference on computer vision and pattern recognition}, pages 770--778, 2016.

\bibitem[Heinzerling and Strube(2017)]{heinzerling2017bpemb}
Benjamin Heinzerling and Michael Strube.
\newblock Bpemb: Tokenization-free pre-trained subword embeddings in 275 languages.
\newblock \emph{arXiv preprint arXiv:1710.02187}, 2017.

\bibitem[Hsu and Chang(2006)]{hsu06crfcheck}
Y.-F. Hsu and S.-F. Chang.
\newblock Detecting image splicing using geometry invariants and camera characteristics consistency.
\newblock In \emph{International Conference on Multimedia and Expo}, 2006.

\bibitem[Hu et~al.(2018)Hu, Shen, and Sun]{hu2018squeeze}
Jie Hu, Li Shen, and Gang Sun.
\newblock Squeeze-and-excitation networks.
\newblock In \emph{Proceedings of the IEEE conference on computer vision and pattern recognition}, pages 7132--7141, 2018.

\bibitem[Hu et~al.(2020)Hu, Zhang, Jiang, Chaudhuri, Yang, and Nevatia]{hu2020span}
Xuefeng Hu, Zhihan Zhang, Zhenye Jiang, Syomantak Chaudhuri, Zhenheng Yang, and Ram Nevatia.
\newblock Span: Spatial pyramid attention network for image manipulation localization.
\newblock In \emph{Computer Vision--ECCV 2020: 16th European Conference, Glasgow, UK, August 23--28, 2020, Proceedings, Part XXI 16}, pages 312--328. Springer, 2020.

\bibitem[Huh et~al.(2018)Huh, Liu, Owens, and Efros]{huh2018fighting}
Minyoung Huh, Andrew Liu, Andrew Owens, and Alexei~A Efros.
\newblock Fighting fake news: Image splice detection via learned self-consistency.
\newblock In \emph{Proceedings of the European conference on computer vision (ECCV)}, pages 101--117, 2018.

\bibitem[Kadam et~al.(2021)Kadam, Ahirrao, and Kotecha]{kadam2021multiple}
Kalyani~Dhananjay Kadam, Swati Ahirrao, and Ketan Kotecha.
\newblock Multiple image splicing dataset (misd): a dataset for multiple splicing.
\newblock \emph{Data}, 6\penalty0 (10):\penalty0 102, 2021.

\bibitem[Kanopoulos et~al.(1988)Kanopoulos, Vasanthavada, and Baker]{kanopoulos1988design}
Nick Kanopoulos, Nagesh Vasanthavada, and Robert~L Baker.
\newblock Design of an image edge detection filter using the sobel operator.
\newblock \emph{IEEE Journal of solid-state circuits}, 23\penalty0 (2):\penalty0 358--367, 1988.

\bibitem[Kawar et~al.(2023)Kawar, Zada, Lang, Tov, Chang, Dekel, Mosseri, and Irani]{kawar2023imagic}
Bahjat Kawar, Shiran Zada, Oran Lang, Omer Tov, Huiwen Chang, Tali Dekel, Inbar Mosseri, and Michal Irani.
\newblock Imagic: Text-based real image editing with diffusion models.
\newblock In \emph{Proceedings of the IEEE/CVF Conference on Computer Vision and Pattern Recognition}, pages 6007--6017, 2023.

\bibitem[Kingma and Ba(2014)]{kingma2014adam}
Diederik~P Kingma and Jimmy Ba.
\newblock Adam: A method for stochastic optimization.
\newblock \emph{arXiv preprint arXiv:1412.6980}, 2014.

\bibitem[Koenderink(1984)]{koenderink1984structure}
Jan~J Koenderink.
\newblock The structure of images.
\newblock \emph{Biological cybernetics}, 50\penalty0 (5):\penalty0 363--370, 1984.

\bibitem[Korus and Huang(2016)]{korus2016multi}
Pawe{\l} Korus and Jiwu Huang.
\newblock Multi-scale analysis strategies in prnu-based tampering localization.
\newblock \emph{IEEE Transactions on Information Forensics and Security}, 12\penalty0 (4):\penalty0 809--824, 2016.

\bibitem[Lee et~al.(2024)Lee, Kang, and Han]{lee2024diffusion}
Hyunsoo Lee, Minsoo Kang, and Bohyung Han.
\newblock Diffusion-based conditional image editing through optimized inference with guidance.
\newblock \emph{arXiv preprint arXiv:2412.15798}, 2024.

\bibitem[Li and Shen(2022)]{li2022wavesnet}
Qiufu Li and Linlin Shen.
\newblock Wavesnet: Wavelet integrated deep networks for image segmentation.
\newblock In \emph{Chinese Conference on Pattern Recognition and Computer Vision (PRCV)}, pages 325--337. Springer, 2022.

\bibitem[Li et~al.(2020)Li, Shen, Guo, and Lai]{li2020wavelet}
Qiufu Li, Linlin Shen, Sheng Guo, and Zhihui Lai.
\newblock Wavelet integrated cnns for noise-robust image classification.
\newblock In \emph{Proceedings of the IEEE/CVF conference on computer vision and pattern recognition}, pages 7245--7254, 2020.

\bibitem[Lin et~al.(2023)Lin, Tan, Xu, Ma, and Lau]{lin2023frequency}
Jiaying Lin, Xin Tan, Ke Xu, Lizhuang Ma, and Rynson~WH Lau.
\newblock Frequency-aware camouflaged object detection.
\newblock \emph{ACM Transactions on Multimedia Computing, Communications and Applications}, 19\penalty0 (2):\penalty0 1--16, 2023.

\bibitem[Liu et~al.(2024{\natexlab{a}})Liu, Li, and Ding]{liu2024referring}
Chang Liu, Xiangtai Li, and Henghui Ding.
\newblock Referring image editing: Object-level image editing via referring expressions.
\newblock In \emph{Proceedings of the IEEE/CVF Conference on Computer Vision and Pattern Recognition}, pages 13128--13138, 2024{\natexlab{a}}.

\bibitem[Liu et~al.(2024{\natexlab{b}})Liu, Chen, Peng, Wang, Hu, and Gao]{liu2024attention}
Decheng Liu, Tao Chen, Chunlei Peng, Nannan Wang, Ruimin Hu, and Xinbo Gao.
\newblock Attention consistency refined masked frequency forgery representation for generalizing face forgery detection.
\newblock \emph{IEEE Transactions on Information Forensics and Security}, 2024{\natexlab{b}}.

\bibitem[Liu et~al.(2021{\natexlab{a}})Liu, Li, Zhou, Chen, He, Xue, Zhang, and Yu]{liu2021spatial}
Honggu Liu, Xiaodan Li, Wenbo Zhou, Yuefeng Chen, Yuan He, Hui Xue, Weiming Zhang, and Nenghai Yu.
\newblock Spatial-phase shallow learning: rethinking face forgery detection in frequency domain.
\newblock In \emph{Proceedings of the IEEE/CVF conference on computer vision and pattern recognition}, pages 772--781, 2021{\natexlab{a}}.

\bibitem[Liu et~al.(2022)Liu, Liu, Chen, and Liu]{liu2022pscc}
Xiaohong Liu, Yaojie Liu, Jun Chen, and Xiaoming Liu.
\newblock Pscc-net: Progressive spatio-channel correlation network for image manipulation detection and localization.
\newblock \emph{IEEE Transactions on Circuits and Systems for Video Technology}, 32\penalty0 (11):\penalty0 7505--7517, 2022.

\bibitem[Liu et~al.(2021{\natexlab{b}})Liu, Lin, Cao, Hu, Wei, Zhang, Lin, and Guo]{liu2021swin}
Ze Liu, Yutong Lin, Yue Cao, Han Hu, Yixuan Wei, Zheng Zhang, Stephen Lin, and Baining Guo.
\newblock Swin transformer: Hierarchical vision transformer using shifted windows.
\newblock In \emph{Proceedings of the IEEE/CVF international conference on computer vision}, pages 10012--10022, 2021{\natexlab{b}}.

\bibitem[Loshchilov and Hutter(2016)]{loshchilov2016sgdr}
Ilya Loshchilov and Frank Hutter.
\newblock Sgdr: Stochastic gradient descent with warm restarts.
\newblock \emph{arXiv preprint arXiv:1608.03983}, 2016.

\bibitem[Lowe(2004)]{lowe2004distinctive}
David~G Lowe.
\newblock Distinctive image features from scale-invariant keypoints.
\newblock \emph{International journal of computer vision}, 60:\penalty0 91--110, 2004.

\bibitem[Luo et~al.(2021)Luo, Zhang, Yan, and Liu]{luo2021generalizing}
Yuchen Luo, Yong Zhang, Junchi Yan, and Wei Liu.
\newblock Generalizing face forgery detection with high-frequency features.
\newblock In \emph{Proceedings of the IEEE/CVF conference on computer vision and pattern recognition}, pages 16317--16326, 2021.

\bibitem[Milletari et~al.(2016)Milletari, Navab, and Ahmadi]{milletari2016v}
Fausto Milletari, Nassir Navab, and Seyed-Ahmad Ahmadi.
\newblock V-net: Fully convolutional neural networks for volumetric medical image segmentation.
\newblock In \emph{2016 fourth international conference on 3D vision (3DV)}, pages 565--571. Ieee, 2016.

\bibitem[Mubarak et~al.(2023)Mubarak, Alsboui, Alshaikh, Inuwa-Dutse, Khan, and Parkinson]{mubarak2023survey}
Rami Mubarak, Tariq Alsboui, Omar Alshaikh, Isa Inuwa-Dutse, Saad Khan, and Simon Parkinson.
\newblock A survey on the detection and impacts of deepfakes in visual, audio, and textual formats.
\newblock \emph{Ieee Access}, 11:\penalty0 144497--144529, 2023.

\bibitem[Nam et~al.(2024)Nam, Syazwany, Kim, and Lee]{nam2024modality}
Ju-Hyeon Nam, Nur~Suriza Syazwany, Su~Jung Kim, and Sang-Chul Lee.
\newblock Modality-agnostic domain generalizable medical image segmentation by multi-frequency in multi-scale attention.
\newblock In \emph{Proceedings of the IEEE/CVF Conference on Computer Vision and Pattern Recognition}, pages 11480--11491, 2024.

\bibitem[Niloy et~al.(2023)Niloy, Bhaumik, and Woo]{niloy2023cfl}
Fahim~Faisal Niloy, Kishor~Kumar Bhaumik, and Simon~S Woo.
\newblock Cfl-net: image forgery localization using contrastive learning.
\newblock In \emph{Proceedings of the IEEE/CVF Winter Conference on Applications of Computer Vision}, pages 4642--4651, 2023.

\bibitem[Novozamsky et~al.(2020)Novozamsky, Mahdian, and Saic]{Novozamsky_2020_WACV}
Adam Novozamsky, Babak Mahdian, and Stanislav Saic.
\newblock Imd2020: A large-scale annotated dataset tailored for detecting manipulated images.
\newblock In \emph{2020 IEEE Winter Applications of Computer Vision Workshops (WACVW)}, pages 71--80, 2020.

\bibitem[Pham et~al.(2019)Pham, Lee, Kwon, and Park]{pham2019hybrid}
Nam~Thanh Pham, Jong-Weon Lee, Goo-Rak Kwon, and Chun-Su Park.
\newblock Hybrid image-retrieval method for image-splicing validation.
\newblock \emph{Symmetry}, 11\penalty0 (1):\penalty0 83, 2019.

\bibitem[Popescu and Farid(2004)]{popescu2004statistical}
Alin~C Popescu and Hany Farid.
\newblock Statistical tools for digital forensics.
\newblock In \emph{International workshop on information hiding}, pages 128--147. Springer, 2004.

\bibitem[Qian et~al.(2020)Qian, Yin, Sheng, Chen, and Shao]{qian2020thinking}
Yuyang Qian, Guojun Yin, Lu Sheng, Zixuan Chen, and Jing Shao.
\newblock Thinking in frequency: Face forgery detection by mining frequency-aware clues.
\newblock In \emph{European conference on computer vision}, pages 86--103. Springer, 2020.

\bibitem[Qin et~al.(2021)Qin, Zhang, Wu, and Li]{qin2021fcanet}
Zequn Qin, Pengyi Zhang, Fei Wu, and Xi Li.
\newblock Fcanet: Frequency channel attention networks.
\newblock In \emph{Proceedings of the IEEE/CVF international conference on computer vision}, pages 783--792, 2021.

\bibitem[Redi et~al.(2011)Redi, Taktak, and Dugelay]{redi2011digital}
Judith~A Redi, Wiem Taktak, and Jean-Luc Dugelay.
\newblock Digital image forensics: a booklet for beginners.
\newblock \emph{Multimedia Tools and Applications}, 51:\penalty0 133--162, 2011.

\bibitem[Ronneberger et~al.(2015)Ronneberger, Fischer, and Brox]{ronneberger2015u}
Olaf Ronneberger, Philipp Fischer, and Thomas Brox.
\newblock U-net: Convolutional networks for biomedical image segmentation.
\newblock In \emph{Medical image computing and computer-assisted intervention--MICCAI 2015: 18th international conference, Munich, Germany, October 5-9, 2015, proceedings, part III 18}, pages 234--241. Springer, 2015.

\bibitem[Russakovsky et~al.(2015)Russakovsky, Deng, Su, Krause, Satheesh, Ma, Huang, Karpathy, Khosla, Bernstein, et~al.]{russakovsky2015imagenet}
Olga Russakovsky, Jia Deng, Hao Su, Jonathan Krause, Sanjeev Satheesh, Sean Ma, Zhiheng Huang, Andrej Karpathy, Aditya Khosla, Michael Bernstein, et~al.
\newblock Imagenet large scale visual recognition challenge.
\newblock \emph{International journal of computer vision}, 115:\penalty0 211--252, 2015.

\bibitem[Sang and Hansen(2022)]{sang2022multi}
Mufan Sang and John~HL Hansen.
\newblock Multi-frequency information enhanced channel attention module for speaker representation learning.
\newblock \emph{arXiv preprint arXiv:2207.04540}, 2022.

\bibitem[Tahir and Bal(2024)]{tahir2024deep}
Eren Tahir and Mert Bal.
\newblock Deep image composition meets image forgery, 2024.

\bibitem[Tralic et~al.(2013)Tralic, Zupancic, Grgic, and Grgic]{tralic2013comofod}
Dijana Tralic, Ivan Zupancic, Sonja Grgic, and Mislav Grgic.
\newblock Comofod—new database for copy-move forgery detection.
\newblock In \emph{Proceedings ELMAR-2013}, pages 49--54. IEEE, 2013.

\bibitem[Vaccari and Chadwick(2020)]{vaccari2020deepfakes}
Cristian Vaccari and Andrew Chadwick.
\newblock Deepfakes and disinformation: Exploring the impact of synthetic political video on deception, uncertainty, and trust in news.
\newblock \emph{Social media+ society}, 6\penalty0 (1):\penalty0 2056305120903408, 2020.

\bibitem[Wang et~al.(2022{\natexlab{a}})Wang, Wu, Chen, Han, Shrivastava, Lim, and Jiang]{wang2022objectformer}
Junke Wang, Zuxuan Wu, Jingjing Chen, Xintong Han, Abhinav Shrivastava, Ser-Nam Lim, and Yu-Gang Jiang.
\newblock Objectformer for image manipulation detection and localization.
\newblock In \emph{Proceedings of the IEEE/CVF Conference on Computer Vision and Pattern Recognition}, pages 2364--2373, 2022{\natexlab{a}}.

\bibitem[Wang et~al.(2022{\natexlab{b}})Wang, Xie, Li, Fan, Song, Liang, Lu, Luo, and Shao]{wang2022pvt}
Wenhai Wang, Enze Xie, Xiang Li, Deng-Ping Fan, Kaitao Song, Ding Liang, Tong Lu, Ping Luo, and Ling Shao.
\newblock Pvt v2: Improved baselines with pyramid vision transformer.
\newblock \emph{Computational Visual Media}, 8\penalty0 (3):\penalty0 415--424, 2022{\natexlab{b}}.

\bibitem[Woo et~al.(2018)Woo, Park, Lee, and Kweon]{woo2018cbam}
Sanghyun Woo, Jongchan Park, Joon-Young Lee, and In~So Kweon.
\newblock Cbam: Convolutional block attention module.
\newblock In \emph{Proceedings of the European conference on computer vision (ECCV)}, pages 3--19, 2018.

\bibitem[Wu et~al.(2019)Wu, AbdAlmageed, and Natarajan]{wu2019mantra}
Yue Wu, Wael AbdAlmageed, and Premkumar Natarajan.
\newblock Mantra-net: Manipulation tracing network for detection and localization of image forgeries with anomalous features.
\newblock In \emph{Proceedings of the IEEE/CVF conference on computer vision and pattern recognition}, pages 9543--9552, 2019.

\bibitem[Wu et~al.(2022)Wu, Liu, Zhan, and Cheng]{wu2022p2t}
Yu-Huan Wu, Yun Liu, Xin Zhan, and Ming-Ming Cheng.
\newblock P2t: Pyramid pooling transformer for scene understanding.
\newblock \emph{IEEE transactions on pattern analysis and machine intelligence}, 2022.

\bibitem[Xu et~al.(2024)Xu, Chen, Lv, Wang, and Zhang]{xu2024image}
Xu Xu, Junxin Chen, Wenrui Lv, Wei Wang, and Yushu Zhang.
\newblock Image tampering detection with frequency-aware attention and multi-view fusion.
\newblock \emph{IEEE Transactions on Artificial Intelligence}, 2024.

\bibitem[Yang et~al.(2023)Yang, Gu, Zhang, Zhang, Chen, Sun, Chen, and Wen]{yang2023paint}
Binxin Yang, Shuyang Gu, Bo Zhang, Ting Zhang, Xuejin Chen, Xiaoyan Sun, Dong Chen, and Fang Wen.
\newblock Paint by example: Exemplar-based image editing with diffusion models.
\newblock In \emph{Proceedings of the IEEE/CVF Conference on Computer Vision and Pattern Recognition}, pages 18381--18391, 2023.

\bibitem[Yang et~al.(2020)Yang, Zhang, Zhu, and Kwong]{yang2020clustering}
Jianquan Yang, Yulan Zhang, Guopu Zhu, and Sam Kwong.
\newblock A clustering-based framework for improving the performance of jpeg quantization step estimation.
\newblock \emph{IEEE transactions on circuits and systems for video technology}, 31\penalty0 (4):\penalty0 1661--1672, 2020.

\bibitem[Yang and Soatto(2020)]{yang2020fda}
Yanchao Yang and Stefano Soatto.
\newblock Fda: Fourier domain adaptation for semantic segmentation.
\newblock In \emph{Proceedings of the IEEE/CVF conference on computer vision and pattern recognition}, pages 4085--4095, 2020.

\bibitem[Yang et~al.(2024)Yang, Liu, Bi, Xiao, Li, Wang, and Gao]{yang2024d}
Zonglin Yang, Bo Liu, Xiuli Bi, Bin Xiao, Weisheng Li, Guoyin Wang, and Xinbo Gao.
\newblock D-net: A dual-encoder network for image splicing forgery detection and localization.
\newblock \emph{Pattern Recognition}, 155:\penalty0 110727, 2024.

\bibitem[Yue et~al.(2020)Yue, Robert, and Ungerleider]{yue2020curvature}
Xiaomin Yue, Sophia Robert, and Leslie~G Ungerleider.
\newblock Curvature processing in human visual cortical areas.
\newblock \emph{NeuroImage}, 222:\penalty0 117295, 2020.

\bibitem[Zhang et~al.(2023)Zhang, Jin, Jiang, An, and Lyu]{zhang2023wcanet}
Fukai Zhang, Xiaobo Jin, Jie Jiang, Shan An, and Qiang Lyu.
\newblock Wcanet: Wavelet channel attention network for citrus variety identification.
\newblock In \emph{2023 IEEE International Conference on Image Processing (ICIP)}, pages 2845--2849. IEEE, 2023.

\bibitem[Zhang et~al.(2022)Zhang, Wu, Zhang, Zhu, Lin, Zhang, Sun, He, Mueller, Manmatha, et~al.]{zhang2022resnest}
Hang Zhang, Chongruo Wu, Zhongyue Zhang, Yi Zhu, Haibin Lin, Zhi Zhang, Yue Sun, Tong He, Jonas Mueller, R Manmatha, et~al.
\newblock Resnest: Split-attention networks.
\newblock In \emph{Proceedings of the IEEE/CVF conference on computer vision and pattern recognition}, pages 2736--2746, 2022.

\bibitem[Zhang et~al.(2021)Zhang, Zhu, Wu, Kwong, Zhang, and Zhou]{zhang2021multi}
Yulan Zhang, Guopu Zhu, Ligang Wu, Sam Kwong, Hongli Zhang, and Yicong Zhou.
\newblock Multi-task se-network for image splicing localization.
\newblock \emph{IEEE Transactions on Circuits and Systems for Video Technology}, 32\penalty0 (7):\penalty0 4828--4840, 2021.

\bibitem[Zhong et~al.(2022)Zhong, Li, Tang, Kuang, Wu, and Ding]{zhong2022detecting}
Yijie Zhong, Bo Li, Lv Tang, Senyun Kuang, Shuang Wu, and Shouhong Ding.
\newblock Detecting camouflaged object in frequency domain.
\newblock In \emph{Proceedings of the IEEE/CVF conference on computer vision and pattern recognition}, pages 4504--4513, 2022.

\bibitem[Zhou et~al.(2018)Zhou, Han, Morariu, and Davis]{zhou2018learning}
Peng Zhou, Xintong Han, Vlad~I Morariu, and Larry~S Davis.
\newblock Learning rich features for image manipulation detection.
\newblock In \emph{Proceedings of the IEEE conference on computer vision and pattern recognition}, pages 1053--1061, 2018.

\end{thebibliography}
}
\clearpage
\setcounter{page}{1}
\maketitlesupplementary

\begin{table}[]
    \centering
    \scriptsize
    \begin{tabular}{c|c||c|c}
    \hline
    Dataset Name & Total Images & Copy-Move & Splicing \\
    \hline
    CASIAv2 \cite{pham2019hybrid} & 5,123 & 3,274 & 1,849 \\
    CASIAv1 \cite{Dong2013}       & 920 & 459 & 461 \\
    DIS25k \cite{tahir2024deep}   & 24,964 & 0 & 24,964 \\
    Columbia \cite{hsu06crfcheck} & 180 & 0 & 180 \\
    IMD2020 \cite{Novozamsky_2020_WACV} & 2,010 & - & - \\ 
    CoMoFoD \cite{tralic2013comofod} & 260 & 260 & 0 \\
    In the Wild \cite{huh2018fighting} & 201 & 0 & 201 \\
    MISD \cite{kadam2021multiple} & 300 & 0 & 300 \\
    \hline
    \end{tabular}
    \caption{Summary of the datasets used in this paper.}
    \label{tab:dataset_summary}
\end{table}

\section{Dataset Descriptions}
\label{appendix_dataset_descriptions}

\begin{itemize}
    \item \textbf{CASIAv1} \cite{Dong2013} and \textbf{CASIAv2} \cite{pham2019hybrid}: The CASIAv1 dataset consists of JPG images with a resolution of 384 × 256, including 459 copy move and 461 splicing images. CASIAv2 is more complex than CASIAv1, containing 5,123 tampered ones which consists of 3274 copy move images and 1849 splicing images. The image sizes range from 320 × 240 to 800 × 600 and are available in multiple formats, such as uncompressed BMP and TIFF.

    \item \textbf{DIS25k} \cite{tahir2024deep}: The DIS25k dataset is a large-scale image splicing dataset designed to enhance the realism and complexity of manipulated images. It contains 24,964 spliced images, generated using image composition techniques, such as deep image matting and harmonization, to improve the seamlessness of manipulation. The dataset was created by leveraging the OPA dataset for rational object placement and refining the images with advanced blending techniques, making the forgeries harder to detect. The image sizes vary but generally range from 512 × 512 to 1920 × 1080, ensuring diversity in resolution.

    \item \textbf{Columbia} \cite{hsu06crfcheck}: The Columbia dataset is predominantly altered using splicing and contains high-resolution images. The manipulated regions often span large portions of scenes. Image sizes range from 757 × 568 to 1152 × 768 and are provided in TIFF or BMP formats. In total, it includes 180 tampered images.

    \item \textbf{IMD2020} \cite{Novozamsky_2020_WACV}: The IMD2020 dataset features a diverse set of tampered images collected from real-world sources on the internet, totaling approximately 2,010 manipulated samples. The images are provided in JPG and TIFF formats.

    \item \textbf{CoMoFoD} \cite{tralic2013comofod}: The CoMoFoD dataset is specifically designed for Copy-Move Forgery Detection (CMFD) and comprises 260 forged image sets, divided into two resolution categories: small (512 × 512) and large (3000 × 2000). The images are classified into five groups based on the type of manipulation applied: translation, rotation, scaling, combination, and distortion. Various post-processing techniques, including JPEG compression, blurring, noise addition, and color reduction, are applied to both the tampered and original images.

    \item  \textbf{In the Wild} \cite{huh2018fighting}: The In-the-Wild dataset is a collection of 201 manipulated images gathered from online sources such as THE ONION (a parody news website) and REDDIT PHOTOSHOP BATTLES (an online community focused on image manipulations). These images represent real-world, naturally occurring spliced forgeries. The images in this dataset come in various sizes, reflecting the diversity of online manipulations.

    \item \textbf{MISD} \cite{kadam2021multiple}: The Multiple Image Splicing Dataset (MSID) is the first publicly available dataset specifically designed for multiple image splicing detection. It contains 300 multiple spliced images, all in JPG format with a resolution of 384 × 256. The dataset was created by combining images from the CASIAv1 dataset and applying multiple splicing operations using Figma software. The spliced images feature various post-processing techniques such as rotation and scaling to enhance realism.
\end{itemize}

\begin{table}[t]
    \centering
    \scriptsize
    \begin{tabular}{c|cc}
    \hline
    Method & Parameters (M) & FLOPs (G) \\
    \hline
    UNet \cite{ronneberger2015u}       & 29.9 & 104.3 \\ 
    SegNet \cite{badrinarayanan2017segnet} & 15.7 & 44.3 \\ 
    MantraNet \cite{wu2019mantra} & 3.63 & 135.7 \\
    RRUNet \cite{bi2019rru} & 5.1 & 45.1 \\
    MT-SENet \cite{zhang2021multi} & 10.1 & 34.2 \\
    TransForensic \cite{hao2021transforensics} & 27.6 & 22.4 \\
    MVSSNet \cite{dong2022mvss} & 136.2 & 31.6 \\
    FBINet \cite{gu2022fbi} & 104.4 & 86.5 \\
    SegNeXt \cite{guo2022segnext} & 28.4 & 19.1 \\
    CFLNet \cite{niloy2023cfl} & 49.2 & 52.2 \\
    EITLNet \cite{guo2024effective} & 50.1 & 35.0 \\
    PIMNet \cite{bai2025pim} & 73.3 & 17.9 \\
    \hline
    \textbf{M2SFormer (Ours)} & 27.4 & 14.2 \\ 
    \hline
    \end{tabular}
    \caption{The number of parameters (M), and FLOPs (G) of different models.}
    \label{tab:efficiency_analysis}
\end{table}

\section{Technical Novelty of M2SFormer}
\textbf{M2SFormer} introduces a unified framework that integrates multi-frequency and multi-scale attention in a single stream, addressing a longstanding challenge in forgery localization where frequency- and spatial-domain features were traditionally processed separately. By enriching the encoder–decoder pipeline with the \textit{M2S Attention Block}, it effectively captures both local, fine-grained anomalies and global, context-aware cues within the same network pass. Furthermore, M2SFormer leverages a \textit{Difficulty-guided Attention (DGA) mechanism} that quantifies each sample’s complexity through a curvature-based global prior map, then automatically generates textual difficulty cues to guide the Transformer decoder’s focus on challenging regions. This approach removes the need for extra metadata and provides adaptive attention control, ultimately boosting cross-domain generalization and boundary precision. The design also emphasizes computational efficiency, fusing multi-frequency representations within the feature space rather than the raw input space, thereby maintaining a balance between performance gains and inference speed. Consequently, M2SFormer stands out for its holistic combination of frequency- and spatial-domain attention, adaptive difficulty-based text guidance, and efficient, robust architecture—all contributing to significant improvements in forgery localization across diverse domains.

\section{Broader Impact in Artificial Intelligence}
\label{sec:broader_impact_in_artificial_intelligence}

\textbf{M2SFormer}’s unified approach to forgery localization—combining \textit{multi-spectral and multi-scale attention} with the \textit{text-guided difficulty attention}—not only enhances detection accuracy for unseen or subtle manipulations but also carries significant broader implications for AI. By capturing both spatial and frequency-domain cues, the model strengthens its cross-domain generalization, \textbf{assisting in the fight against misinformation across social media, journalism, and legal contexts}. Its Difficulty-guided Attention mechanism offers a scalable strategy for other AI tasks that require dynamic allocation of computational resources, potentially improving performance in areas like object detection, medical imaging, and fine-grained recognition. Furthermore, the integration of textual cues into a visual framework underscores the promise of cross-modal solutions, paving the way for more holistic approaches to content understanding. \textit{While adversarial refinement of forgery techniques remains a concern, M2SFormer sets a higher bar for detecting tampered content, promoting authenticity, accountability, and ethical AI development}.

\section{More Detailed Ablation Study on M2SFormer}
\label{sec:detailed_ablation_study}
In this section, we perform a more detailed ablation study on M2SFormer.

\subsection{Ablation Study on Backbone in M2SFormer}

\begin{table}[h]
    \centering
    \scriptsize
    \setlength\tabcolsep{2.0pt} 
    \begin{tabular}{c|c|cc|cc|c|c}
    \hline
    Network & \multicolumn{1}{c|}{\multirow{2}{*}{Backbone}} & \multicolumn{2}{c|}{\textit{Seen}}  & \multicolumn{2}{c|}{\textit{Unseen}} & \multicolumn{1}{c|}{\multirow{2}{*}{Param (M)}}  & \multicolumn{1}{c}{\multirow{2}{*}{FLOPs (G)}} \\ \cline{3-6}
    Type    & & DSC & mIoU & DSC & mIoU & & \\
    \hline
    \multicolumn{1}{c|}{\multirow{3}{*}{CNN}} & ResNet50 & 40.3 & 32.7 & 35.6 & 26.3 & 25.7MB & 14.9GB \\
     & Res2Net50 & 49.9 & 42.1 & 38.1 & 29.1 & 25.9MB & 15.9GB \\
     & ResNeSt50 & 51.1 & 43.2 & 38.5 & 29.5 & 27.6MB & 18.0GB \\
     \hline
    \multicolumn{1}{c|}{\multirow{3}{*}{Transformer}} & MiT-B2 & \textcolor{blue}{\textbf{\textit{56.5}}} & \textcolor{blue}{\textbf{\textit{48.7}}} & \textcolor{blue}{\textbf{\textit{41.2}}} & \textcolor{blue}{\textbf{\textit{32.6}}} & 26.2MB & 12.3GB \\
     & P2T-Small & 55.8 & 48.1 & 40.6 & 32.1 & 25.6MB & 13.5GB \\
     & \textbf{PVT-v2-B2 \tiny{(Ours)}} & \textcolor{red}{\textbf{\underline{57.8}}} & \textcolor{red}{\textbf{\underline{50.8}}} & \textcolor{red}{\textbf{\underline{43.0}}} & \textcolor{red}{\textbf{\underline{34.3}}} & 27.4MB & 14.2GB \\
     \hline
    \end{tabular}
    \caption{Quantitative results on \textit{Seen} (CASIAv2 \cite{pham2019hybrid}) and \textit{Unseen} datasets (Other test datasets) according to backbone network. The efficiency metrics are measured at resolution $256 \times 256$.}
    \label{tab:ablation_backbone_networks}
\end{table}

In this section, we conduct an ablation study to evaluate the impact of different backbone models on the performance of M2SFormer. This experiment uses several popular CNN and Transformer architectures, including ResNet50 \cite{he2016deep}, Res2Net \cite{gao2019res2net}, ResNeSt50 \cite{zhang2022resnest}, MiT-B2 \cite{guo2022segnext}, PVT-v2-b2 \cite{wang2022pvt}, and P2T-Small \cite{wu2022p2t}. Notably, only the backbone network was changed, while all other architectural settings remained consistent with those in the main experiment. We reported the \textit{mean} performance of five-fold cross-validation results for each \textit{Seen} (CASIAv2 \cite{pham2019hybrid}) and \textit{Unseen} datasets (Other test datasets) in Tab. \ref{tab:ablation_backbone_networks}. The datasets used in the \textit{seen} and \textit{unseen} datasets are the same as Tab. \ref{tab:casiav2_training_scheme}.

\section{Metrics Descriptions}
\label{appendix_metric_descriptions}

In this section, we describe the metrics used in this paper. For convenience, we denote $TP, FP$, and $FN$ as the number of samples of true positive, false positive, and false negative between two binary masks $A$ and $B$.  

\begin{itemize}
    \item The \textit{Mean Dice Similarity Coefficient (DSC)} \cite{milletari2016v} measures the similarity between two samples and is widely used to assess the performance of segmentation tasks, such as image segmentation or object detection. \textbf{\underline{Higher is better}}. For given two binary masks $A$ and $B$, DSC is defined as follows:
    \begin{equation}
        \textbf{DSC}(A, B) = \frac{2 \times | A \cap B |}{| A \cup B |} = \frac{2 \times TP}{2 \times TP + FP + FN}
    \end{equation}

    \item The \textit{Mean Intersection over Union (mIoU)} measures the ratio of the intersection area to the union area between predicted and ground truth masks in segmentation tasks. \textbf{\underline{Higher is better}}. For given two binary masks $A$ and $B$, mIoU is defined as follows:
    \begin{equation}
        \textbf{mIoU}(A, B) = \frac{A \cap B}{A \cup B} = \frac{TP}{TP + FP + FN}
    \end{equation}
\end{itemize}

\begin{figure*}[t]
    \centering
    \includegraphics[width=\textwidth]{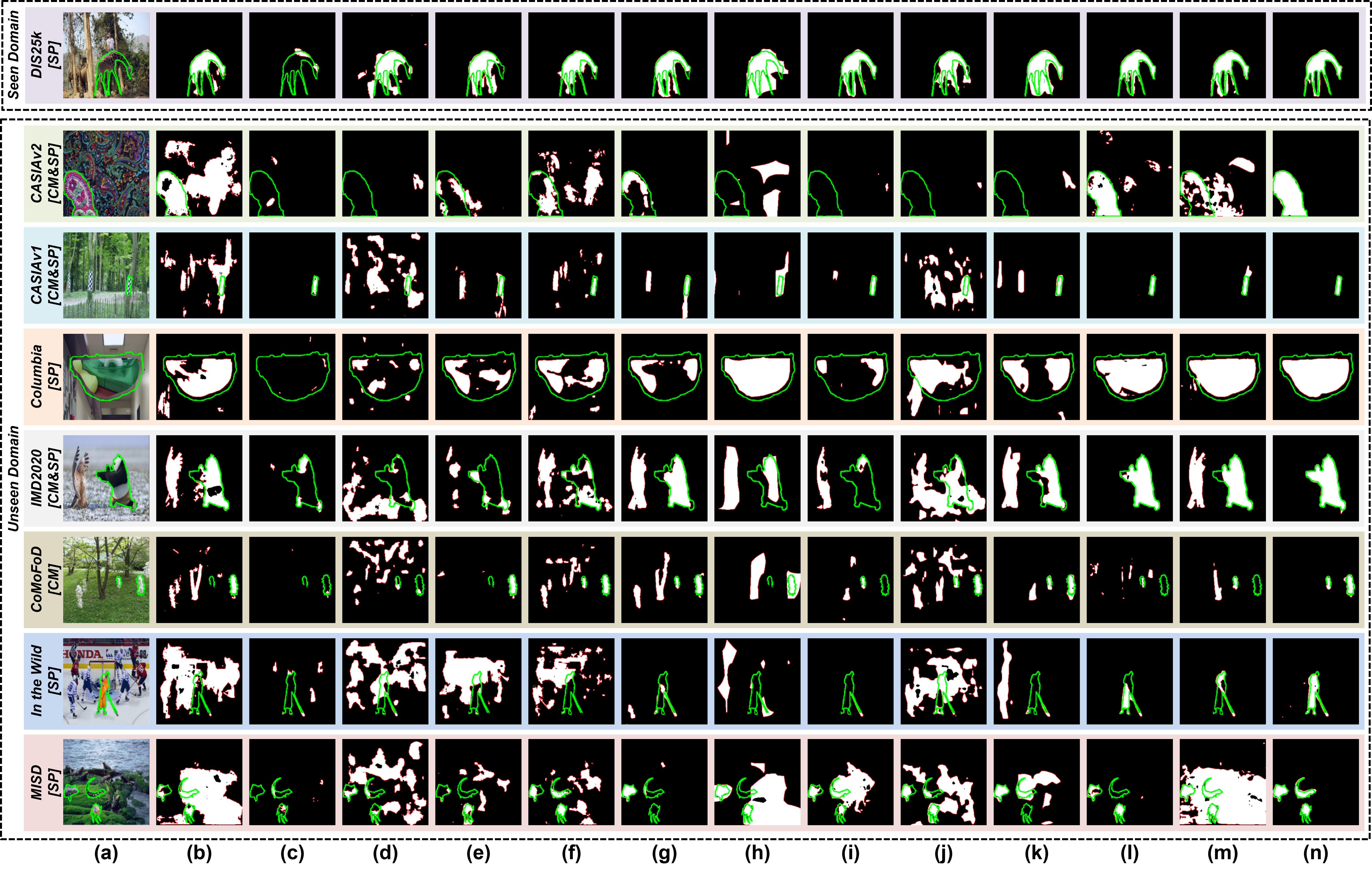}
        \caption{Qualitative comparison of other methods and M2SFormer with DIS25k training scheme.  (a) Input images with ground truth. (b) UNet \cite{ronneberger2015u}. (c) SegNet \cite{badrinarayanan2017segnet}. (d) MantraNet \cite{wu2019mantra}. (e) RRUNet \cite{bi2019rru}. (f) MT-SENet \cite{zhang2021multi}. (g) TransForensic \cite{hao2021transforensics}. (h) MVSSNet \cite{dong2022mvss}. (i) FBINet \cite{gu2022fbi}. (j) SegNeXt \cite{guo2022segnext}. (k) CFLNet \cite{niloy2023cfl}. (l) EITLNet \cite{guo2024effective}. (m) PIMNet \cite{bai2025pim}. (n) \textbf{M2SFormer (Ours)}. \textcolor{green}{\textbf{Green}} and \textcolor{red}{\textbf{Red}} lines denote the boundaries of the ground truth and prediction, respectively. And, the “SP” and “CM” in square brackets represent the types of forgery in the dataset: “SP” stands for Splicing, while “CM” denotes Copy-Move.}
    \label{fig:SupQualitativeResults}
\end{figure*}

\end{document}